\documentclass[letterpaper, 10 pt, conference]{ieeeconf}  

\IEEEoverridecommandlockouts                              

\usepackage{amsmath} 
\usepackage{amssymb}  

\usepackage[backend=biber,style=ieee,natbib=true,maxnames=2,minnames=1,maxbibnames=99]{biblatex}
\usepackage{booktabs}
\usepackage{graphicx}
\usepackage[footnotesize]{caption}
\usepackage{subfig}
\usepackage{tcolorbox}
\usepackage{gensymb}
\usepackage{hyperref}
\usepackage{balance}
\hypersetup{
breaklinks=true,letterpaper=true,colorlinks,linkcolor=[RGB]{153,27,30},citecolor=[RGB]{153,27,30},urlcolor=[RGB]{153,27,30}
    }

\usepackage[compact]{titlesec}
\titlespacing{\section}{0pt}{3px}{3px}
\titlespacing{\subsection}{0pt}{2px}{2px}
\titlespacing{\subsubsection}{0pt}{2px}{1px}
\expandafter\def\expandafter\normalsize\expandafter{%
    \normalsize%
    \setlength\abovedisplayskip{5pt}%
    \setlength\belowdisplayskip{5pt}%
    \setlength\abovedisplayshortskip{0pt}%
    \setlength\belowdisplayshortskip{0pt}%
}

\title{\LARGE \bf
GABRIL: Gaze-Based Regularization for\\Mitigating Causal Confusion in Imitation Learning
}

\author{Amin Banayeeanzade$^{*}$, Fatemeh Bahrani$^{*}$, Yutai Zhou, Erdem Bıyık 
\thanks{* Equal contribution}
\thanks{All authors are with Thomas Lord Department of Computer Science, University of Southern California, USA.}\thanks{Emails: \small \tt{\{banayeea,fb\_269,yutaizho,biyik\}@usc.edu}}%
}

\begin{document}

\maketitle
\thispagestyle{empty}
\pagestyle{empty}


\begin{abstract}
Imitation Learning (IL) is a widely adopted approach which enables agents to learn from human expert demonstrations by framing the task as a supervised learning problem. However, IL often suffers from \textit{causal confusion}, where agents misinterpret spurious correlations as causal relationships, leading to poor performance in testing environments with distribution shift. To address this issue, we introduce GAze-Based Regularization in Imitation Learning (GABRIL), a novel method that leverages the human gaze data gathered during the data collection phase to guide the representation learning in IL. GABRIL utilizes a regularization loss which encourages the model to focus on causally relevant features identified through expert gaze and consequently mitigates the effects of confounding variables. We validate our approach in Atari environments and the Bench2Drive benchmark in CARLA by collecting human gaze datasets and applying our method in both domains. Experimental results show that the improvement of GABRIL over behavior cloning is around $179\%$ more than the same number for other baselines in the Atari and $76\%$ in the CARLA setup. Finally, we show that our method provides extra explainability when compared to regular IL agents. The datasets, the code, and some experiment videos are publicly available at \url{https://liralab.usc.edu/gabril}.
\end{abstract}

\section{Introduction}

Imitation learning (IL) is a framework for training agents by learning from expert demonstrations, enabling them to replicate desired behaviors in a given environment. A commonly used approach in IL is behavior cloning (BC), where an agent learns a policy by mapping observations to actions through supervised learning on expert data \cite{ALVINN}. While effective in controlled settings, BC often suffers from \textit{causal confusion}—the agent learns spurious correlations instead of true causal relationships. As a result, it learns to misattribute the features from the demonstrations that are predictive of the desired actions but are not actually causal \cite{Causal_Confusion_in_IL}.

\begin{figure}[t] 
  \centering
  \includegraphics[width=0.95\linewidth]{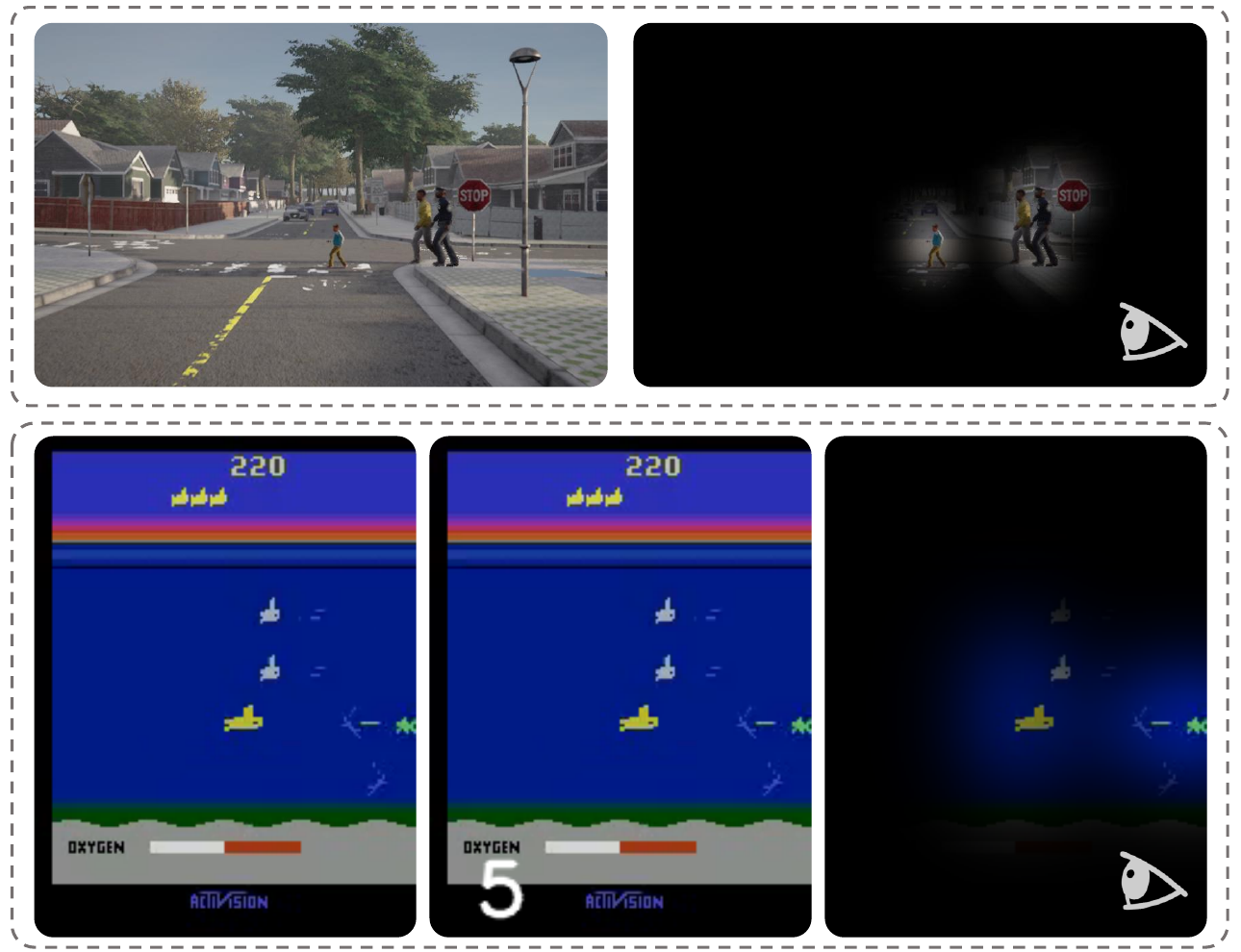}
  
  \caption{Demonstrations from our datasets. \textbf{(Top)} A human expert waits by a stop sign for pedestrians to pass. While there are many misleading visual factors in the environment, the expert can provide extra supervision through their gaze to limit the model's attention to the true underlying factors in the environment. {\textbf{(Bottom-Left)}} A frame of the Seaquest Atari game. {\textbf{(Middle)}} The \textit{confounded} Atari environment in which adding a previous action indicator to the training images considerably misleads the agent. {\textbf{(Right)}} However, a human operator only focuses on the causally relevant factors and ignores the confounding distractors.
  } \label{fig:intro_figure}
  \vspace{-6mm}
\end{figure}

As an example inspired by \citet{Causal_Confusion_in_IL}, consider a self-driving vehicle in an urban environment with frequent stoplights. The correct causal factor for braking is the state of the traffic light. However, if the demonstration images come from within the vehicle's cab with a visible brake light indicator on the dashboard, the agent may erroneously learn to rely on a shortcut—to brake only when the light is on rather than identifying the true causal factor. This is because all of the braking instances in the training data will occur when the brake light is on. 
As another example, artificially appending the previous action labels to the training images in an Atari game environment might lead to catastrophic failure during testing because the agent merely learns to repeat the recorded action instead of correctly solving the task \cite{Causal_Confusion_in_IL} (see Fig.~\ref{fig:intro_figure}). These examples illustrate how BC methods suffer from causal confusion, as the agent lacks sufficient cues to discern the underlying basis of human expert decisions.

This issue can be addressed if we can identify the factors that influence the expert's actions—that is, what they focus on when making decisions in the environment. While explicitly querying the expert can provide this information \cite{Causal_Confusion_in_IL}, doing so at every timestep is prohibitively expensive. However, studies in neuroscience show that humans utilize their gaze to selectively attend to causally relevant information for efficiently processing and understanding complex visual scenes \cite {Itti1998Saliency, Darby2021Attention}. Our insight in this paper is that the gaze is a freely available data source as humans naturally use it, and it contains valuable information about the causal factors. By using eye-tracking equipment to record the human gaze, we can access this critical information without requiring additional effort from the expert and further utilize it to address the causal confusion problem.

In this paper, we propose a novel approach called GAze-Based Regularization in Imitation Learning (GABRIL) which mitigates the causal confusion problem by improving representation learning in the latent space of models. More precisely, given an image observation, GABRIL encourages the model to learn a representation with high activations over the causally related factors—identified through the expert’s gaze—while reducing activations in unrelated spatial regions. In this way, GABRIL improves robustness against confounding variables and helps the agent avoid causal confusion.

To show the effectiveness of our method, we conduct experiments in Atari environments in addition to a more realistic benchmark, namely Bench2Drive \cite{bench2drive} developed in CARLA \cite{CARLA}. We collect a dataset consisting of 1,160 minutes of human experts playing Atari games and another dataset of 71 minutes of expert driving in CARLA, both with recorded gaze data. We compare GABRIL with previous approaches and show that while other methods provide improvements over BC, our method achieves $179\%$ in Atari and $76\%$ more improvement in CARLA compared to the next best method, proving to be state-of-the-art in those benchmarks. We also release both datasets alongside our code.

Moreover, we show that our approach is data-efficient and performs well even with a limited amount of gaze data. Finally, we show that models trained using GABRIL are highly interpretable and can provide human-understandable explanations for their decisions, a desired feature in future autonomous agents. 

\section{Related Work} \label{sec:related_work}

\subsection{Causal Confusion} 
Causal confusion, where agents mistakenly attend to irrelevant and non-causal factors for decision-making, is a common problem in training neural networks \cite{clrl}. More specifically, this issue is studied in reinforcement learning \cite{seeing_is_not_believing, causal_deconfounding_drl} and imitation learning \cite{resolve_copycat, Causal_Confusion_in_IL, oreo, CGL}. It is critical to address this problem to improve the reliability and performance of BC agents \cite{survey_end_to_end_autonomous}. 

On this line, \citet{Causal_Confusion_in_IL} highlight how IL models can latch onto spurious correlations rather than true causal relationships, leading to poor generalization. To mitigate this issue, they propose an intervention-based approach that explicitly forces models to learn causal dependencies by leveraging counterfactual data augmentation. Additionally, \textit{OREO} \cite{oreo} is a dropout-based approach that extracts semantic objects with a VQ-VAE \cite{vqvae} and then randomly masks out all pixels contributing to each object, i.e., the units with the same discrete code. Moreover, \citet{SEMI_cc} combine semantic bird-eye views with OREO and propose a sequential masking method to reduce causal confusion. 

Compared to this line of work, GABRIL uses a novel regularization method that employs the human gaze to address the causal confusion problem in BC agents. Furthermore, GABRIL complements the prior work we mentioned here. As an example, we will show in Section~\ref{sec:experiments} that it improves OREO when used in combination.

\subsection{Gaze for Imitation Learning} 
Several prior works have utilized human gaze to enhance BC. For instance, \textit{GRIL} \cite{GRIL} utilizes a neural network to predict both action and gaze coordinates simultaneously using a multi-objective loss in drone-controlling tasks. Gaze information has also been proven effective in robot manipulation tasks \cite{gaze_robot_manipulation_1, gaze_robot_manipulation_2,visarl}, and several studies illustrate improvements in autonomous driving and gaming tasks when benefiting from gaze supervision \cite{biswas2024, GMD, CGL, selective_gaze, AGIL}. 

More specifically, \textit{GMD} \cite{GMD} uses the human gaze to modulate the probability distribution of dropout in 2D activation maps. \citet{selective_gaze} propose an auxiliary model with three modules. The first module predicts the human gaze mask, while the gating module specifies whether the predicted gaze map should be used in learning. If activated by the gating module, the gaze mask is concatenated to the input and fed to the final module for decision-making. Relatedly, \textit{AGIL} \cite{AGIL} offers a model with two parallel processing pathways. The first one takes the raw image, while the second uses the filtered version of the image, which is constructed by multiplying the input by the gaze mask. The outputs of both paths are averaged and fed to the action prediction layers. 

Prior studies have used gaze supervision to specifically address causal confusion. For example, \textit{CGL} \cite{CGL} defines a KL-based coverage metric that penalizes the model if it ignores the regions where the human attends. More related to our work, \citet{biswas2024} present a \textit{contrastive} approach that encourages the model to attend more to relevant areas of the image with a triplet loss: they create positive and negative variants of the input image, such that in the positive variant, the non-causal factors are blurred out, while in the negative image, they blur the causal factors. Using this generated pair of modified images together with the original image as the anchor, their contrastive loss leads to learning representations that encourage the model to attend to the causal factors. 

With a similar motivation, we propose to use gaze as an extra supervision to solve the causal confusion problem. Unlike prior works, our method applies a regularization loss in the latent space to improve the representation learning. In Section~\ref{sec:experiments}, we compare GABRIL against the aforementioned baselines to show the effectiveness of our method. It is also noteworthy that our method does not require gaze information during test time, as opposed to some of the prior methods, e.g., GMD and AGIL.

\section{Preliminaries} 

In this section, we use the formulation of structural causal model (SCM) \cite{peters2017elements} to explain the causal confusion problem in imitation learning, and to later propose our methodology using gaze supervision. In imitation learning, the goal is to train a policy $\pi:\mathcal{O}\to \mathcal{A}$ that predicts an action given an observation $o \in \mathcal{O}$, where $\mathcal{O}$ is the set of all possible observations, e.g., images from the environment, and $\mathcal{A}$ is the space of valid actions. 

In any dynamical system, two sets of underlying factors contribute to observation generation: (i) the set of \textit{causal} factors $\mathcal{S}$, is the set of all variables in the environment that are essential for an optimal agent to make a decision, e.g., the traffic light in the self-driving agent example, and (ii) the set of \textit{confounding} variables $\mathcal{T}$, which consists of all other factors which might be correlated but not causally related in making the correct decision, e.g., the state of the brake light in the cab, the shape of nearby buildings, etc. An ideal policy has the property that a distribution shift in the confounding variables does not affect its output. More formally, we expect the policy $\pi$ to satisfy the following robustness property.

\newtheorem{definition}{Definition}

\definition A policy $\pi$ is robust to the confounding variables if it satisfies
${P(\pi(o) \mid \text{do}(\mathcal{S} = s, \mathcal{T} = t))} = {P(\pi(o) \mid \text{do}(\mathcal{S} = s, \mathcal{T} = t'))}$ for any set of variable assignments $s$, $t$ and $t'$. \label{definition}

The \textit{do} operator here denotes an intervention in the SCM, meaning that we actively set the variables $\mathcal{S}$ and $\mathcal{T}$ to specific values rather than passively observing them \cite{peters2017elements}. In the rest of the paper, we use the notation $\text{do}(s, t)$ instead of ${\text{do}(\mathcal{S} = s, \mathcal{T} = t)}$ for brevity.

In regular BC, the policy \( \pi \) is trained to mimic the behavior of an expert by leveraging a dataset of expert demonstrations. The data collection process involves an expert interacting with the environment to generate a set of $N$ observation-action pairs \(\mathcal{D} = \{(o_i, a_i)\}_{i=1}^{N} \), where each \( o_i \in \mathcal{O}\) represents an observation image and \( a_i \in \mathcal{A}\) is the corresponding action taken by the expert. Formally, BC frames the problem as a supervised learning task in which the objective is to minimize the discrepancy between the policy’s predicted actions and the expert's demonstrated actions. The standard approach is to minimize the empirical risk over the dataset: 
\[
    \mathcal{L}_{\text{BC}}(\pi) =  \frac{1}{N} \sum_{(o_i,a_i) \in \mathcal{D}} \ell(\pi(o_i), a_i),
\]
where \( \ell \) is a suitable loss function, typically the cross-entropy loss for discrete action spaces or mean squared error for continuous action spaces. 

A key challenge in BC, or more broadly supervised learning, is the potential reliance on spurious correlations in training data. Since the policy is trained merely on expert demonstrations, there is no explicit mechanism that enforces it to rely only on causal factors and not confounding variables. As a result, it may inadvertently learn statistical shortcuts that fail to generalize under distribution shifts—an issue commonly discussed in the context of \textit{shortcut learning} or \textit{causal representation learning} \cite{geirhos2020shortcut}.
In what comes next, we introduce our methodology that incorporates gaze supervision to guide the policy toward leveraging causal factors rather than confounders.

\section{Method}
\label{sec:method}

Human experts naturally direct their eye gaze toward causally relevant aspects of the environment \cite{Itti1998Saliency, Darby2021Attention}, providing an additional source of information beyond actions. To utilize this information, we assume access to an input dataset \({\mathcal{D} = \{(o_i, g_i, a_i)\}_{i=1}^{N} }\), where $g_i$ is a saliency mask derived from the human gaze corresponding to the observation $o_i$. We present the details of gaze mask generation in the next subsection and provide our approach to utilizing them in Section~\ref{sec:analysis}.

\begin{figure}[t]
    \centering
        \subfloat[]{
        \includegraphics[width=0.222\columnwidth]{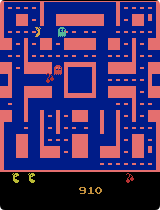}
        } 
        \subfloat[]{
        \includegraphics[width=0.222\columnwidth]{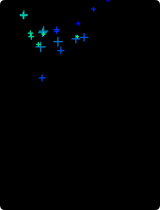}
        }
        \subfloat[]{
        \includegraphics[width=0.222\columnwidth]{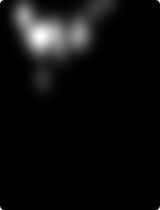}
        }
        \subfloat[]{
        \includegraphics[width=0.222\columnwidth]{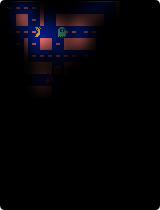}
        }
    \caption{An illustration of the steps involved in building the gaze mask: 
    \textbf{(a)}~An observation $o_i$ of the \textit{MsPacman} Atari game. 
    \textbf{(b)}~The gaze point coordinates $(x_i,y_i)$ corresponding to multiple frames are overlaid as cross markers over a black background. The scale of the marks correlates with the timestep of the collected gaze sample such that the largest cross sign corresponds to the gaze at the present timestep, while smaller ones are either from the future or the past. The color transition from blue to green indicates the progression of time. 
    \textbf{(c)}~The corresponding multimodal mask $g_i$ generated by accumulating the gaze samples.
    \textbf{(d)}~Applying the gaze mask to the raw image to filter out confounding factors that are not essential in decision-making.
    }
    \label{fig:atari_with_gaze}
    \vspace{-5mm}
\end{figure}

\subsection{Gaze Pre-Processing}

When collecting the human expert demonstrations, we record the eye gaze corresponding to each frame of observation as a tuple of coordinates $(x_i,y_i)$ in our dataset. In the next step, we convert the gaze coordinates to a gaze mask by considering the gaze center as the mean of a 2D Gaussian distribution \cite{AGIL} and a manually adjusted variance to account for both the measurement errors and the size of the human fovea, i.e., how much of the image a participant has in focus during a fixation \cite{Bylinskii2019WhatDo}. 

When interacting with a complex environment, humans use their visual short-term memory to retain information about previously attended regions \cite{cowan2001magical}, allowing them to make decisions based on past observations rather than just their current gaze coordinates. Consequently, we build a multimodal Gaussian mask that includes gaze data from both the current and the previous frames when modeling human attention. However, since previously attended objects may have moved or lost relevance over time, we reduce their intensity and expand their Gaussian radius to reflect the natural decay of visual memory while preserving their influence on decision-making. Additionally, 
we found it useful to go beyond and include not only the gaze data from the past but also from the future, as the gaze is likely to land on soon-to-be-important objects in subsequent frames \cite{hollingworth2008understanding}. This is possible because the training happens offline and the gaze data is used only during training. Overall, we construct the multimodal mask by incorporating the gaze data from the present as well as the past and future timesteps with attenuated strengths, which is formulated as:
\begin{align}
    \bar{g}_i = \sum_{j=-k}^{k} {\alpha}^{|j|}\mathcal{N}([x_{i+j}, y_{i+j}], \gamma^2 \beta^{-2|j|}I)
    \label{eq:gaze_mask}
\end{align}
where $\alpha, \beta < 1$ and $\gamma$ are environment specific hyperparameters and $I$ is the identity matrix. Finally, we normalize $\bar{g}_i$ into the range $[0,1]$ to get the gaze mask $g_i$. This multimodal construction over timesteps ensures that fixations gain more weight, as they persist across multiple frames, reinforcing their importance in the attention model. In contrast, saccades, which are rapid transitions between fixations \cite{fixation_saccade}, tend to disappear since they occur over very short durations and do not consistently contribute to any single frame. Fig.~\ref{fig:atari_with_gaze} demonstrates a visualization of the gaze mask generation. The gaze mask generated in this step will be used in the subsequent sections to reduce the causal confusion problem.

\subsection{Analysis} \label{sec:analysis}
Our approach leverages the gaze mask as a supervisory signal to guide BC agents, helping them focus on critical regions and improving their ability to generalize by reducing reliance on misleading correlations. For analysis, we make the following assumption.

\newtheorem{assumption}{Assumption}

\assumption{The gaze $g$, collected from the human expert, is invariant to the changes in confounding variables, i.e., ${P(g \mid \text{do}(s, t))} = {P(g \mid \text{do}(s, t'))}$ for any $s$, $t$, and $t'$.}\label{assumption1}

With this assumption, we focus on the ideal case where the expert's gaze depends only on the causal factors and is not affected by the confounders.
GABRIL uses this invariance property to enhance BC, improving robustness against confounding variables.

To achieve this, we model $\pi(o) = f(\psi(o))$, where $\psi$ is the image encoder, e.g., a convolutional neural network, and $f$ is the action predictor network, e.g., an MLP layer on top of $\psi$. Moreover, we introduce a gaze generator function, $\phi$, which operates on the image encoder $\psi$ to extract the gaze data from the image encoding. Put another way, we consider an ideal image encoder such that $g \approx \phi(\psi(o))$. The following proposition provides sufficient conditions for achieving a robust policy.

\newtheorem{prop}{Proposition} 

\prop \label{prop1} Under Assumption~\ref{assumption1} and an invertible gaze generator function $\phi$, such that $g = \phi(\psi(o))$, the policy $\pi$ is robust to the confounding variables.

\proof{ Consider an arbitrary function $h(g)$ over saliency masks. For any $s$, $t$ and $t'$ it holds that
\begin{align*}
P(h(g) \mid \text{do}(s, t)) = & \int_g P(h(g) \mid g) p(g \mid \text{do}(s, t))
\\ = & \int_g P(h(g) \mid g) p(g \mid \text{do}(s, t'))
\\ = & P(h(g) \mid \text{do}(s, t'))
\end{align*} 
where we used Assumption~\ref{assumption1} in the second equation. Finally, since $\phi$ is assumed to be invertible and $g = \phi(\psi(o))$, we have $\pi = f(\phi^{-1}(g))$. Therefore, substituting $h(\cdot)$ with $f(\phi^{-1}(\cdot))$ in the above equations gives
\begin{align*}
P(\pi(o) \mid \text{do}(s, t)) &= P(f(\phi^{-1}(g)) \mid \text{do}(s, t))\\
&= P(f(\phi^{-1}(g)) \mid \text{do}(s, t'))\\
&= P(\pi(o) \mid \text{do}(s, t')). 
\tag*{\QEDopen}
\end{align*} 

\subsection{Gaze-Based Regularization}

Inspired by Proposition~\ref{prop1}, we propose the following gaze-based regularization loss:
\begin{equation}\label{eq:loss_gp}
\mathcal{L}_{\text{GP}}(\pi) = \frac{1}{NM} \sum_{(o_i,g_i) \in \mathcal{D}}  \Vert \phi(\psi(o_i)) - g_i \Vert_F^2\:,    
\end{equation}
where \( \|\cdot\|_F \) is the Frobenius norm, $N$ is the number of gaze-observation pairs and $M$ is the dimensionality of $g_i$.

In practice, there exist several considerations to apply this loss function. First, Assumption~\ref{assumption1} does not always hold, as even a human expert may get distracted by the confounders. Moreover, even if Assumption~\ref{assumption1} held, achieving a merely robust policy is not our ultimate goal. For example, a constant action predictor $f(\cdot) = 0$ leads to a policy that is robust under Definition~\ref{definition}, but is not what we expect. Similarly, an image encoder $\psi$ that encodes all gaze-related information but discards all the remaining important information for action selection will lead to a policy that is robust to confounders but will fail to produce good actions. 

Instead, we desire a policy that is not only predictive of the gaze but also provides useful features for intelligently selecting the next action in the environment. This requires the image encoder $\psi(o)$ to encode more features than what we need for mere gaze prediction, which means $\phi$ will not be invertible in practice. Therefore, we implement a reasonable, but not necessarily invertible, gaze predictor $\phi$, as we will discuss in the next subsection. Overall, our proposed loss function is
\[
\mathcal{L(\pi)} = \mathcal{L}_{\text{BC}}(\pi) + \lambda \mathcal{L}_{\text{GP}}(\pi)
\]
where $\lambda$ is a hyperparameter to balance the relative strength of the two loss terms.

\begin{figure}[t] 
  \centering
  \vspace{+2mm}
  \includegraphics[width=0.98\columnwidth]{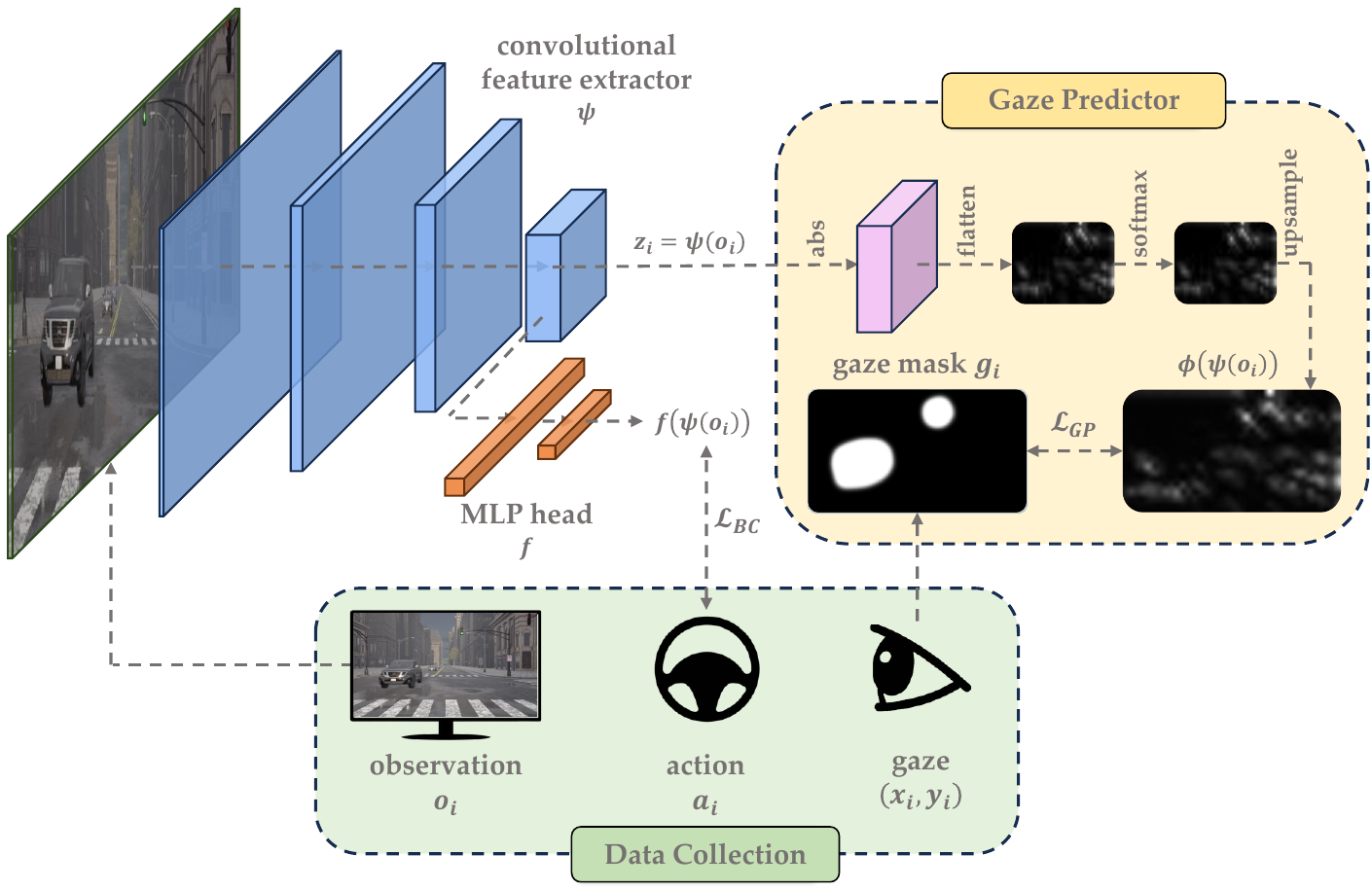}
  \caption{An overview of our proposed method. $\mathcal{L}_{BC}$ is a standard behavioral cloning loss used to reduce the error of action prediction. Besides, we propose a new loss, $\mathcal{L}_{GP}$, which improves the representation learning by regularizing the activation map of the final convolution to follow the gaze mask collected from the human expert. 
  } \label{fig:method}
  \vspace{-7mm}
\end{figure}

\subsection{Gaze-Predictor Design}\label{sec:gaze_pred}

So far, we have discussed the rationale behind incorporating gaze-based regularization into our loss function. Here, we describe the design of the gaze predictor $\phi$ which operates on top of the image encoder $\psi$ to extract the gaze masks. 

Prior studies suggest that the activity levels in the spatial activation map from the final convolutional layer of an image encoder indicate the regions in the input image that the model attends to \cite{Laskin2020Reinforcement}. Consequently, overlaying this map onto the input image enhances interpretability by revealing the model’s focus when making decisions \cite{oreo}. Moreover, knowledge distillation \cite{hinton2015distilling} proves to be particularly effective at this stage when injecting a learned representation of a teacher network into a student \cite{zagoruyko2017paying}. We build upon this line of research by aligning this activation map with the gaze mask—encouraging higher activations in regions attended by humans and lower activations in unattended areas. This approach can be interpreted as a knowledge distillation process in which we align the model's attention with the human gaze pattern.

To achieve this goal, consider an image encoder ${\psi:\mathbb{R}^{c \times h \times w} \to \mathbb{R}^{c' \times h' \times w'}}$, e.g., a convolutional neural network which takes an input image observation with height $h$, width $w$ and $c$ channels, and generates an output activation map with new dimensions $h'$, $w'$ and $c'$, where $h'$ and $w'$ are potentially smaller than the original input dimensions because of the pooling layers and convolutional strides. Let $z = \psi(o)$ denote the activation output of the encoder for input observation $o$.

The gaze mask $g \in \mathbb{R}^{h \times w}$, on the other hand, is a continuous mask in the range $[0,1]$ with the same spatial scale as the input observation. Thus, to construct $g$ from $z$, we perform the following steps, as suggested by prior studies \cite{zagoruyko2017paying, Laskin2020Reinforcement, oreo}. First, we take the absolute value of $z$ followed by mean-pooling across the channel dimension to identify the activations with the highest positive or negative activities. This yields a 2D activation map, over which we apply a softmax. Finally, we upsample the result to match the spatial dimensions of the input image. This procedure defines our proposed gaze predictor $\phi$ on top of vision encoder $\psi$, which is further incorporated in loss function \eqref{eq:loss_gp} to mitigate causal confusion (see Fig.~\ref{fig:method} for a visualization). In Section~\ref{sec:experiments}, we demonstrate the effectiveness of our approach and show how our gaze mask generation method can also enhance explainability during test-time.

\section{Dataset}

In this section, we elaborate on our data collection procedure and the details of the subsequent post-processing steps. First, we utilized the well-known Atari environments implemented in Gymnasium~\cite{gymnasium} and gathered human demonstrations as well as corresponding eye-tracking data. To evaluate our method in a more complex setting, we additionally used the CARLA \cite{CARLA} simulator to train agents for autonomous driving. During data collection, the users were instructed to look only at the monitor screen. Any episodes in which the user was distracted or performed suboptimally were discarded to ensure all demonstrations achieved a 100\% completion rate without collisions in driving and a performance better than the best RL policies in Atari \cite{Causal_Confusion_in_IL}. Fig.~\ref{fig:gaze_setup} shows our experimental setup and provides additional technical details on gaze data collection.

\subsection{Atari Environment}
Our Atari dataset consists of 15 Atari games played for 1,160 minutes in total. 
Each game was rendered at a frame rate convenient for the player, ranging from 10 to 20 FPS. While the player viewed the game at its full-screen resolution, we recorded observations as grayscale images downscaled to ${84\times84}$. The recordings also contain the corresponding gaze data and discrete controller actions for each observation. All games were played with frame skip 4 and a sticky action probability of 0.25 \cite{oreo}. We used $\alpha=0.7$, $\beta=0.99$ and $\gamma=15$ as gaze mask hyperparameters.\footnote{Prior to our work, \citet{zhang2020atari} released an Atari dataset with gaze data. However, our preliminary experiments showed low regular BC performance on that dataset, possibly suggesting suboptimal expert performance. Thus, we collect and release our own dataset.}

\subsection{Self-Driving Environment}
We used the open-source CARLA 0.9.15 \cite{CARLA} to simulate urban driving. Specifically, we used Leaderboard 2.0 framework \cite{leaderboard_v2} to execute scenarios and record data. We leveraged the recently proposed benchmark, Bench2Drive \cite{bench2drive}, comprising 44 driving tasks in different towns and weather conditions. Among all, we selected a diverse subset of 10 driving tasks with the highest potential of having causal confusion and collected 20 expert demonstrations with continuous actions for each task. The recordings contain ${320 \times 180}$ RGB images from a front view camera with ${\text{fov}=60\degree}$, in addition to the collected gaze coordinates, and the continuous action, including brake, steering angle, and throttle. The gaze mask hyperparameters for this environment are adjusted as $\alpha=0.8$, $\beta=0.99$, and $\gamma=30$.

\begin{figure}[t] 
  \centering
  \vspace{+2mm}
  \includegraphics[width=0.9\linewidth]{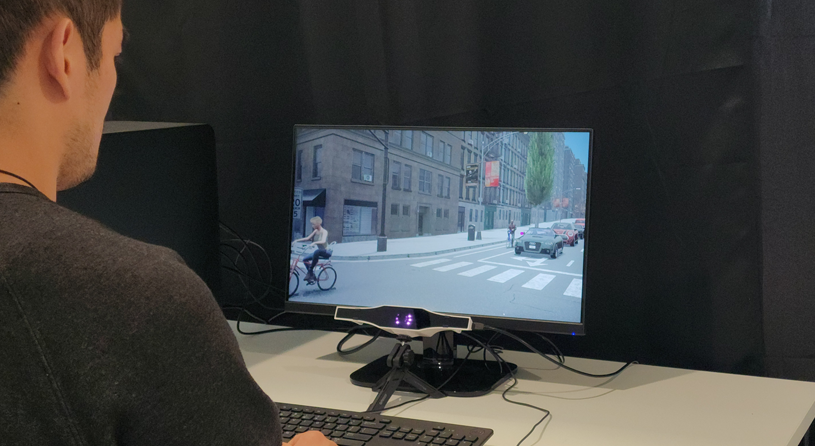}
  \caption{Our data collection setup. The \textit{\href{https://www.gazept.com/product/gp3hd}{GP3 HD}} gaze tracking device is mounted under the monitor, recording the gaze data of the user performing a driving task in the CARLA simulator. 
  The monitor size is 23.8 inches, and the user's distance to the monitor is 65 cm. The gaze tracking device is tilted with an angle of $15\degree$. 
  } \label{fig:gaze_setup}
  \vspace{-5mm}
\end{figure}

\section{Experiments} \label{sec:experiments}

To demonstrate the effectiveness of our proposed method, GABRIL, we empirically investigate the following questions:

\begin{itemize}
    \item How successful is our method relative to other comparable baselines that seek to overcome causal confusion in Atari games? (Tables~\ref{tbl:main_normal} and~\ref{tbl:main_confounded})

    \item How does our method scale with the amount of available gaze and data? (Fig.~\ref{fig:exp_gaze_ratio})

    \item Is our method effective in more realistic environments like CARLA with high-dimensional observations? (Fig.~\ref{fig:carla})

    \item Does our method encode causally relevant features, and how can we leverage it to provide interpretability of the agent decisions? (Fig. \ref{fig_explainability})
    
\end{itemize}

\subsection{Baselines and Metrics}
To answer these questions, we compare our method against several baselines introduced in Section~\ref{sec:related_work} in addition to regular BC. Additionally, we compare against \textit{ViSaRL}~\cite{visarl}, which appends the gaze saliency map to the raw image as an additional input channel and passes the result to the model. 

In our experiments, hyperparameters—such as learning rate, stack size, and regularization coefficients for our method and baselines—were optimized through a hyperparameter search. Specifically, we use $\lambda = 10$ for Atari and $\lambda = {1\mathrm{e}{-4}}$ in CARLA as the gaze loss coefficient for GABRIL. Moreover, to consistently compare the performance of different methods across all Atari games, we use the Advantage over BC (ABC) metric, which quantifies the improvement of each method over the BC baseline, defined as
$$
\text{ABC(M, G)} = \frac{\text{Score(M, G)} - \text{Score(BC, G)}}{\text{Score(BC, G)}},$$
where $\text{Score(M, G)}$ is the score of method $\text{M}$ in the game $\text{G}$. For driving experiments, we adopt the driving score~\cite{bench2drive} (in percentage), which encapsulates a wide range of evaluation metrics, including route completion, collisions, etc. 

\begin{figure}[t] 
    \centering
    \subfloat[DemonAttack]{%
        \label{fig:gaze_ratio_a}
        \includegraphics[width=0.48\columnwidth]{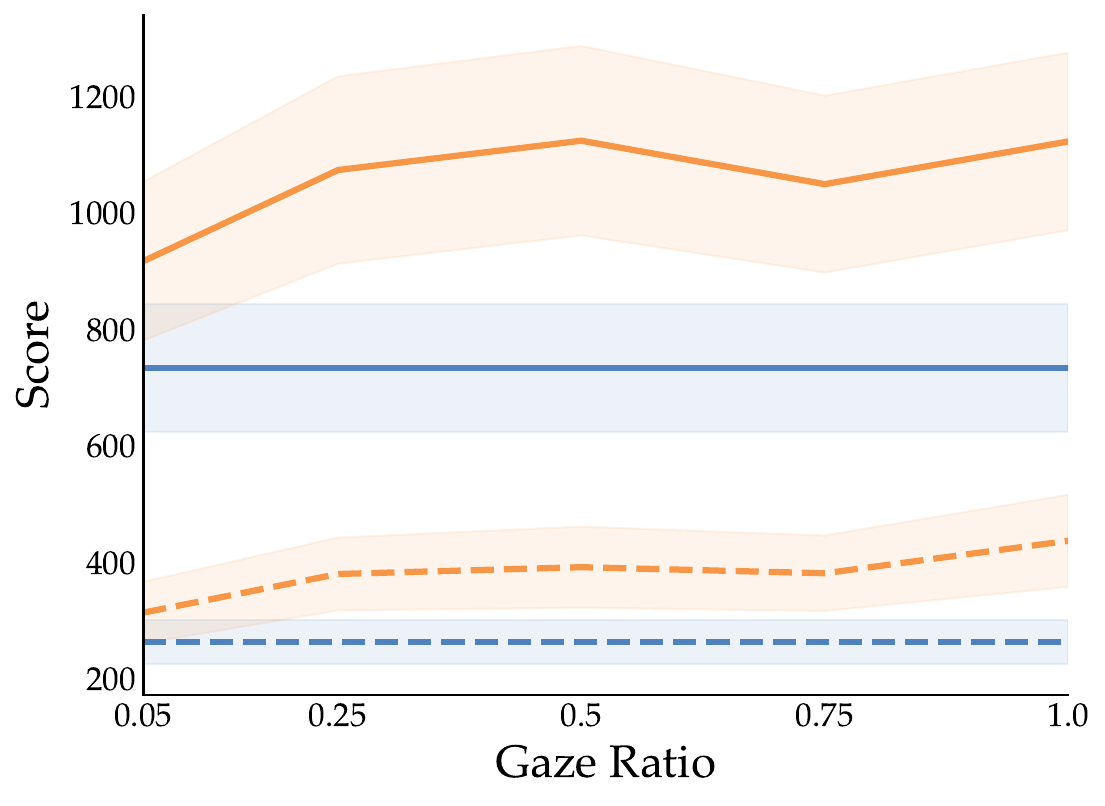}
    }
    \subfloat[Seaquest]{%
       \label{fig:gaze_ratio_b}
        \includegraphics[width=0.48\columnwidth]{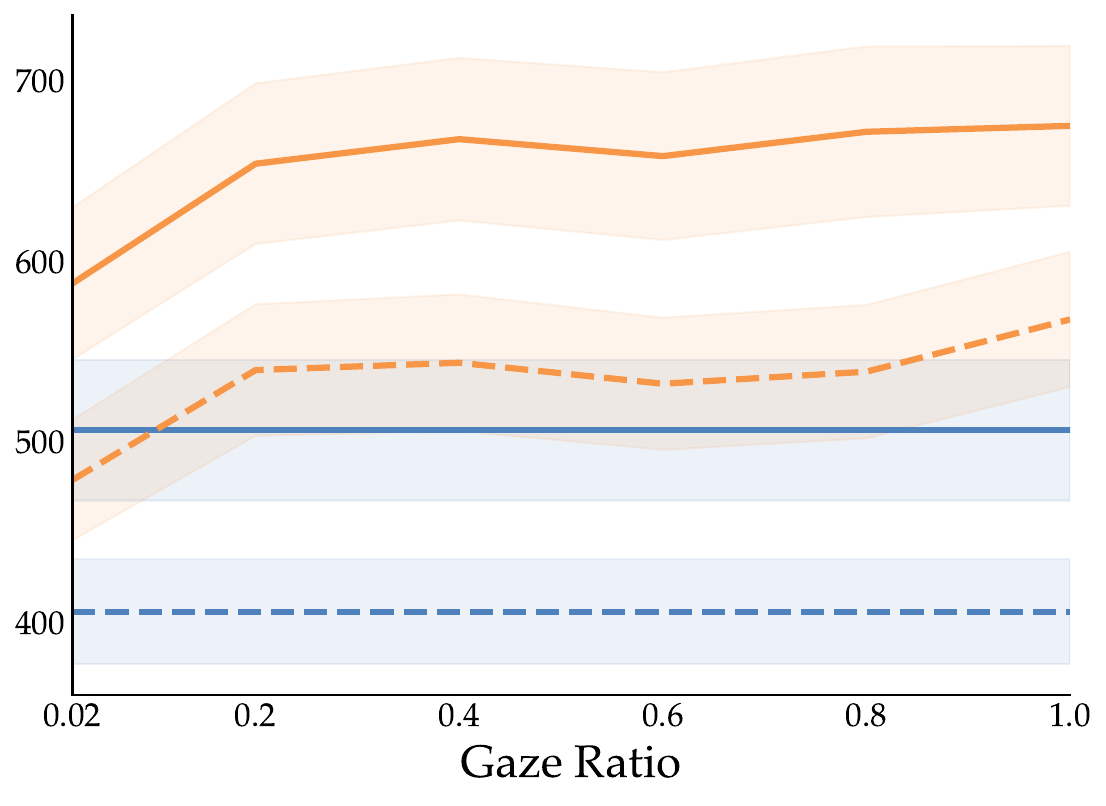}
    }
    
    \vspace{-3mm}
    \subfloat[DemonAttack]{%
       \label{fig:gaze_ratio_c}
        \includegraphics[width=0.48\columnwidth]{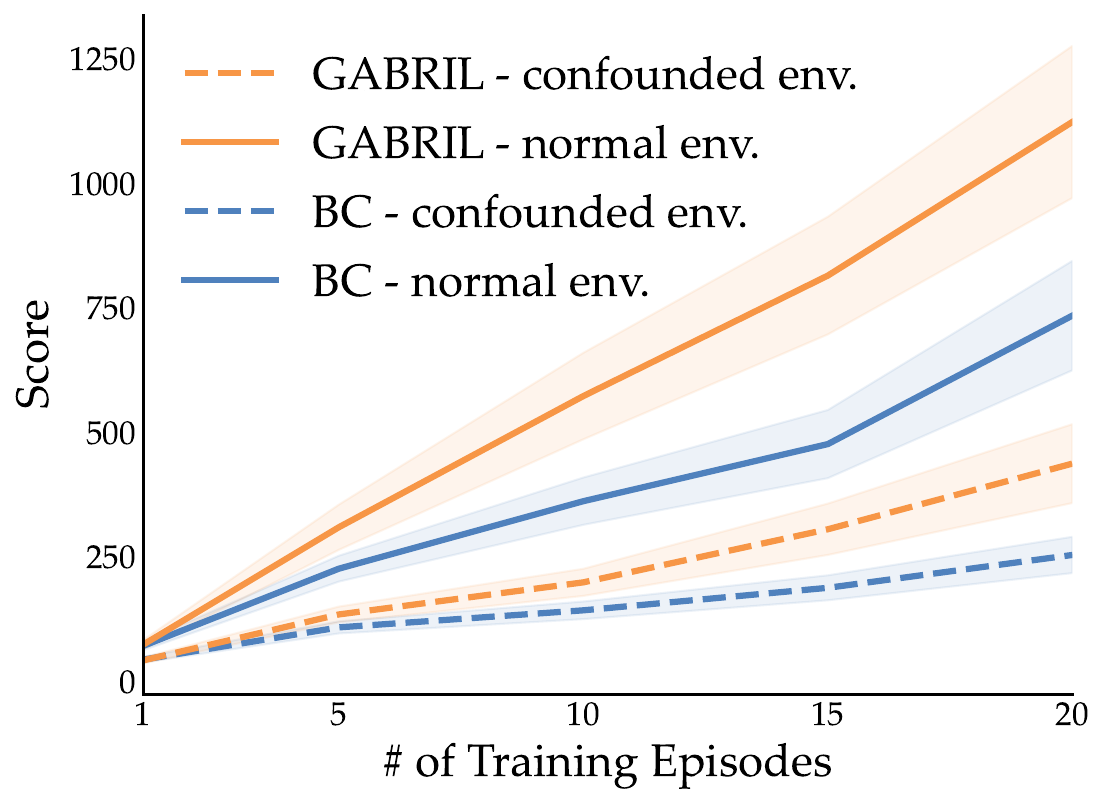}
    }
    \subfloat[Seaquest]{\label{fig:gaze_ratio_d}
    \includegraphics[width=0.48\columnwidth]{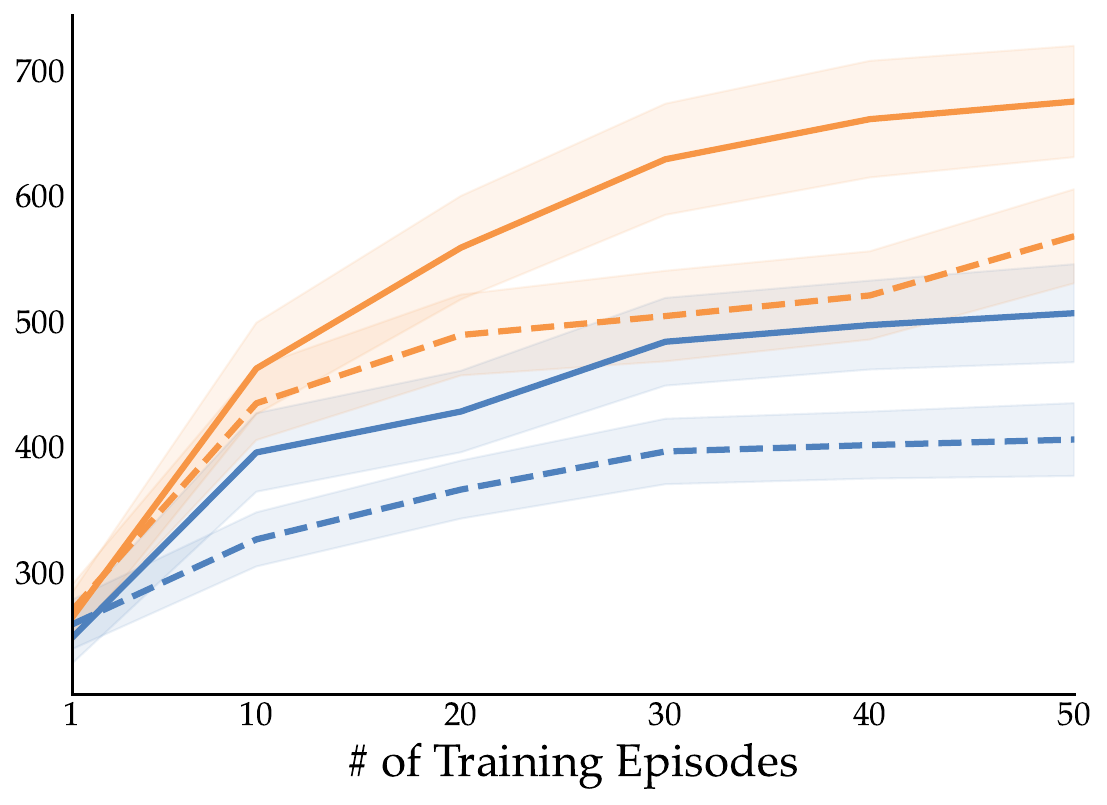}
    }
    \vspace{-3px}
    \caption{Studying the scalability with the amount of data in two Atari environments, \textit{Seaquest} and \textit{DemonAttack}. (a) and (b) show the experiment in which the models use all the expert demonstrations for training, but only a limited portion of the gaze data is included. 
    (c) and (d) show the scalability of our method as a function of the number of (gaze-included) training episodes. The standard deviations are divided by 5 for better visualization.}
    \label{fig:exp_gaze_ratio}
    \vspace{-7mm}
\end{figure}

\subsection{Results}
\noindent\textbf{Atari.} We used two variations of the Atari environments: normal and confounded versions as suggested by \citet{Causal_Confusion_in_IL} (see Fig.~\ref{fig:intro_figure}). For every environment-baseline pair, we trained 8 separate models with different seeds and evaluated each with 100 seeds. We report the mean scores across all trials in Tables~\ref{tbl:main_normal} and~\ref{tbl:main_confounded}. Note that there is no single method that outperforms others consistently in all games, as this was also the case in the work by \citet{oreo}. However, GABRIL successfully outperforms prior regularization methods in the ABC metric, as reported in the last two rows of the tables. Moreover, when combined with dropout methods, our method considerably boosts their performance. For instance, the composition of our method with GMD achieves $22.7\%$ in normal and $32.7\%$ mean ABC in confounded environments, which is near a double improvement when compared to the best baseline. The full version of both tables with the standard deviations is available on the project website.

\noindent\textbf{Data Efficiency.}
Human gaze data may be costly to collect. We investigate if our method has an acceptable performance while only a portion of the gaze data is used. This simulates a use case where a designer may want to use existing gaze-free datasets in conjunction with our algorithm. For this, we fed all the observation-action pairs for \textit{DemonAttack} and \textit{Seaquest} Atari games to the respective models but used only a fraction of the gaze data to apply the regularization loss. Results shown in Figs.~\ref{fig:gaze_ratio_a} and~\ref{fig:gaze_ratio_b} indicate that the model performs reasonably well even with only 20\% of the gaze data. This means we can improve GABRIL's performance by increasing the data while keeping the cost of human gaze collection low. Meanwhile, as depicted in Figs.~\ref{fig:gaze_ratio_c} and~\ref{fig:gaze_ratio_d}, GABRIL is also scalable as the number of gaze-included demonstrations increases.

\begin{figure}[t] 
  \centering
  \vspace{+2mm}
  \includegraphics[width=.85\columnwidth]{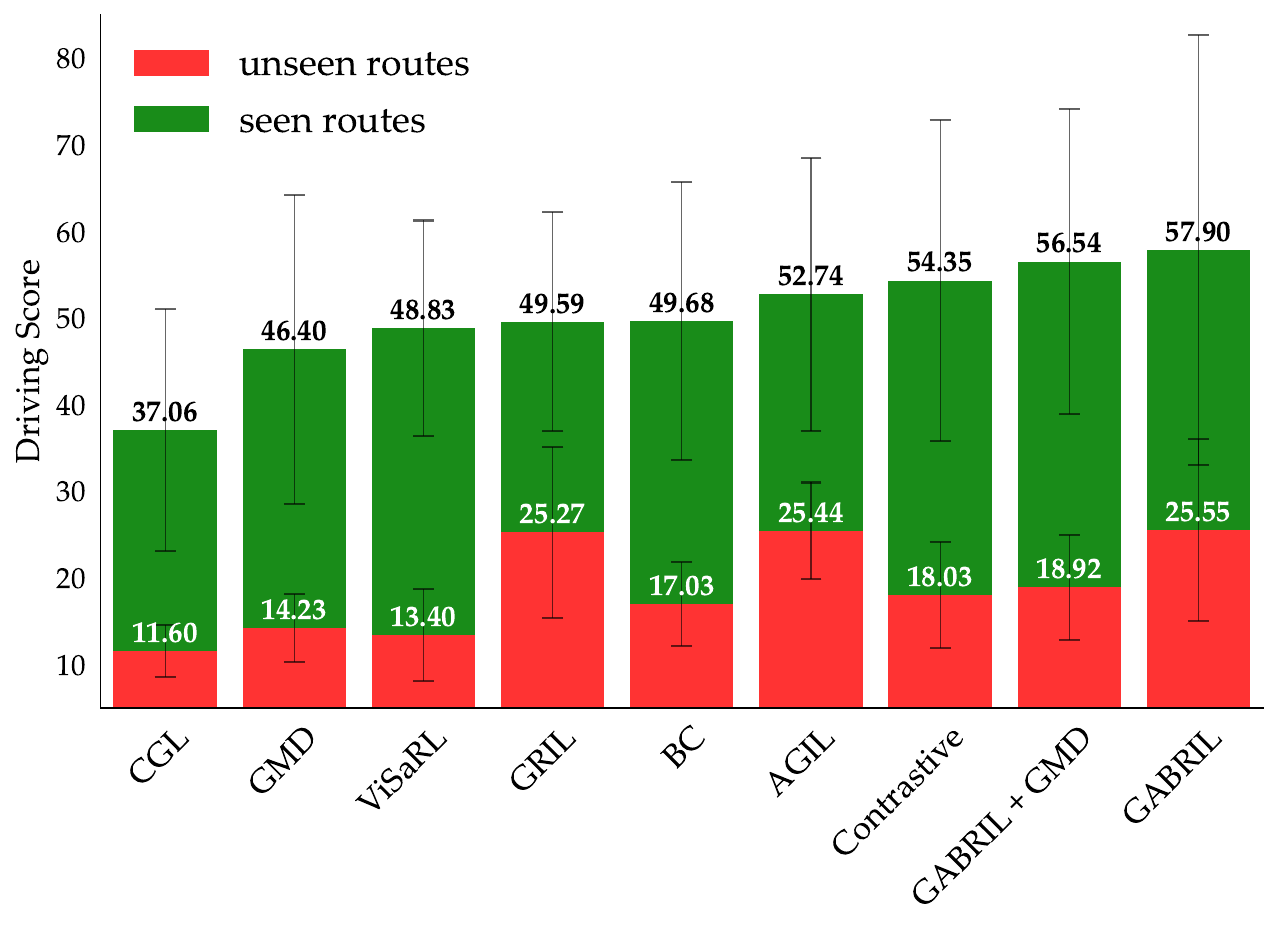}
  \vspace{-2mm}
  \caption{Comparing the performance of different methods in CARLA.
  } \label{fig:carla}
  \vspace{-7mm}
\end{figure}

\begin{figure*}[t]
    \centering
            \centering
            \subfloat{%
                \includegraphics[width=0.24\textwidth]{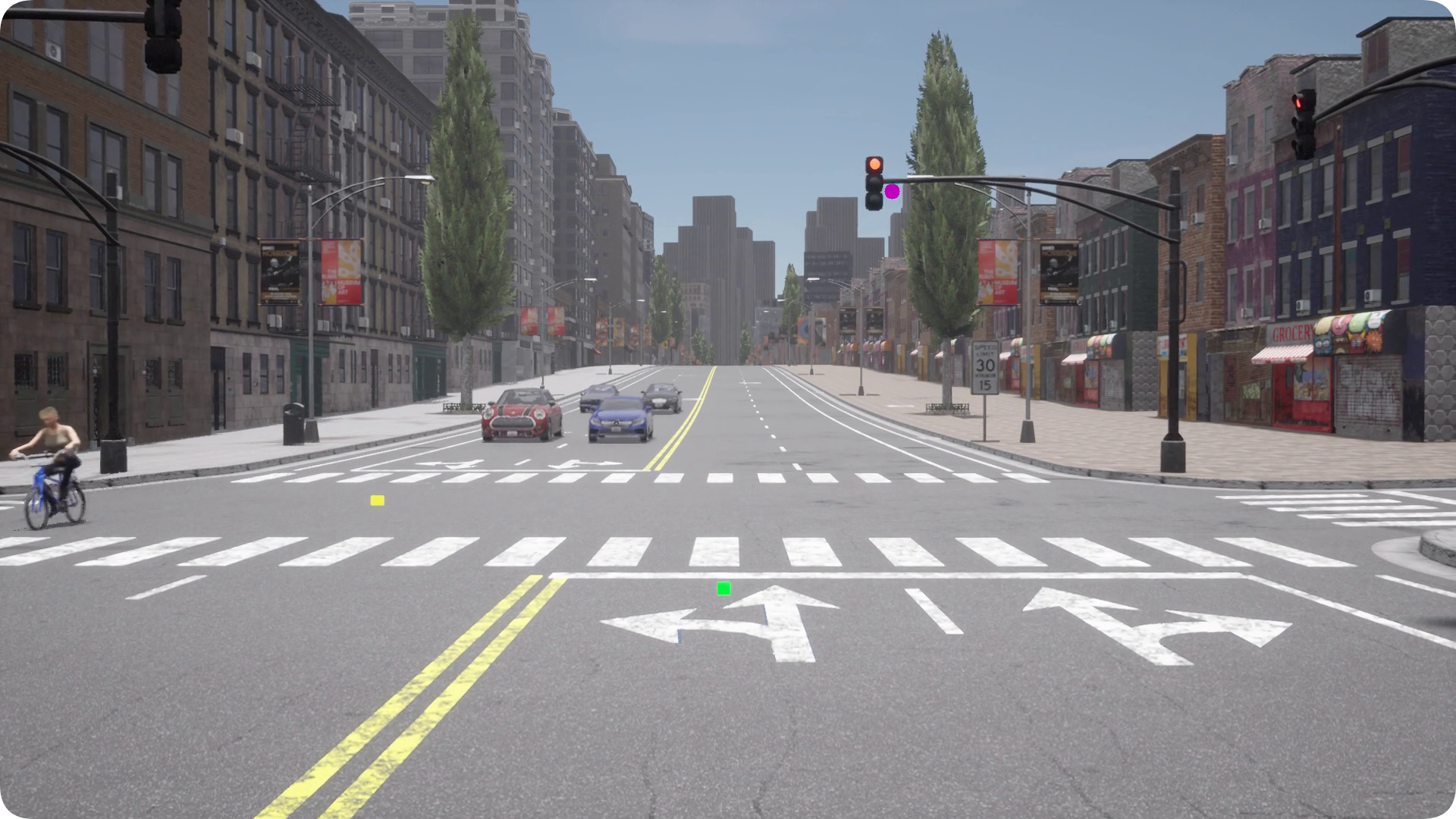}
            }
            \subfloat{%
                \includegraphics[width=0.24\textwidth]{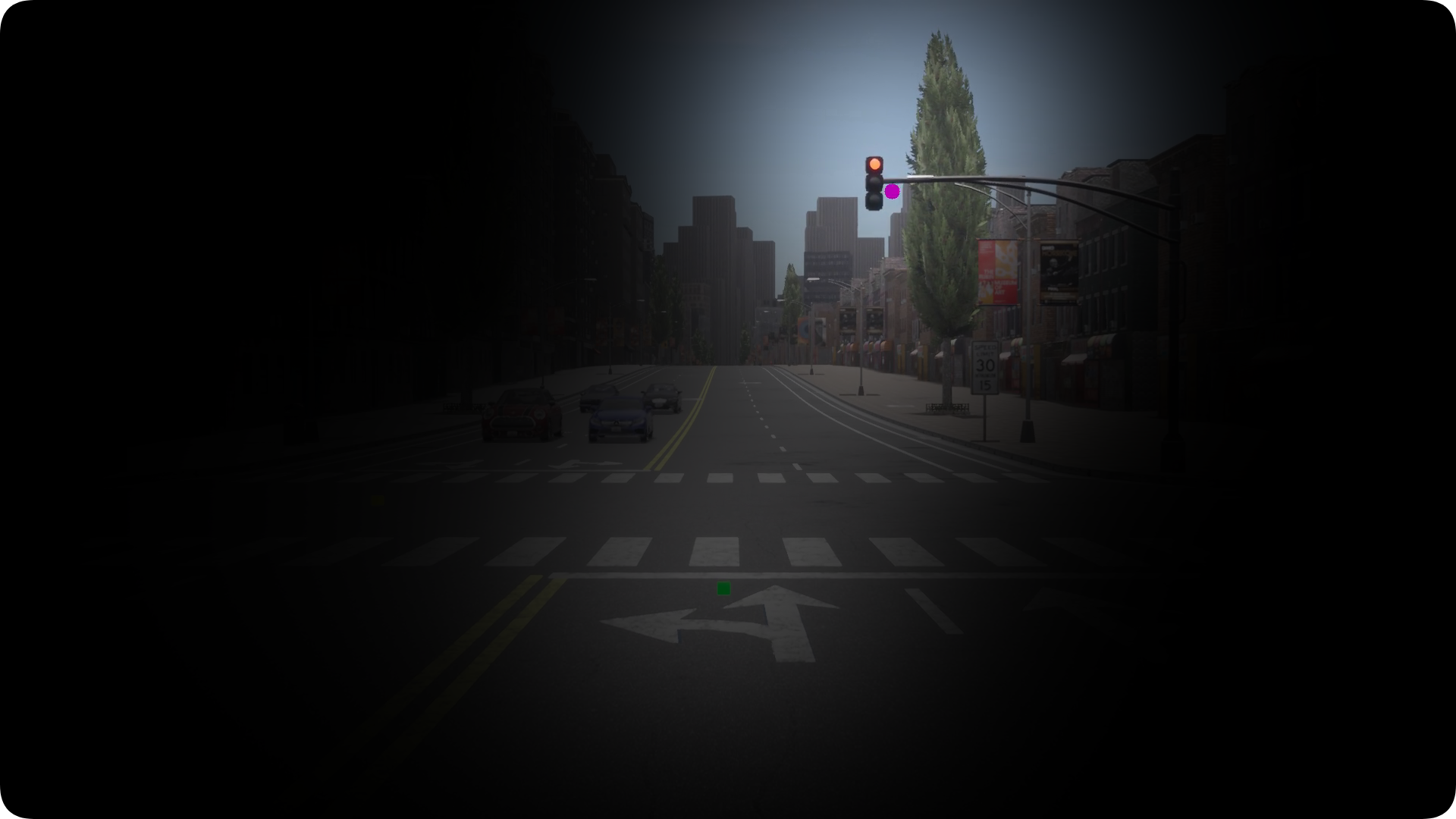}
            }
            \subfloat{%
                \includegraphics[width=0.24\textwidth]{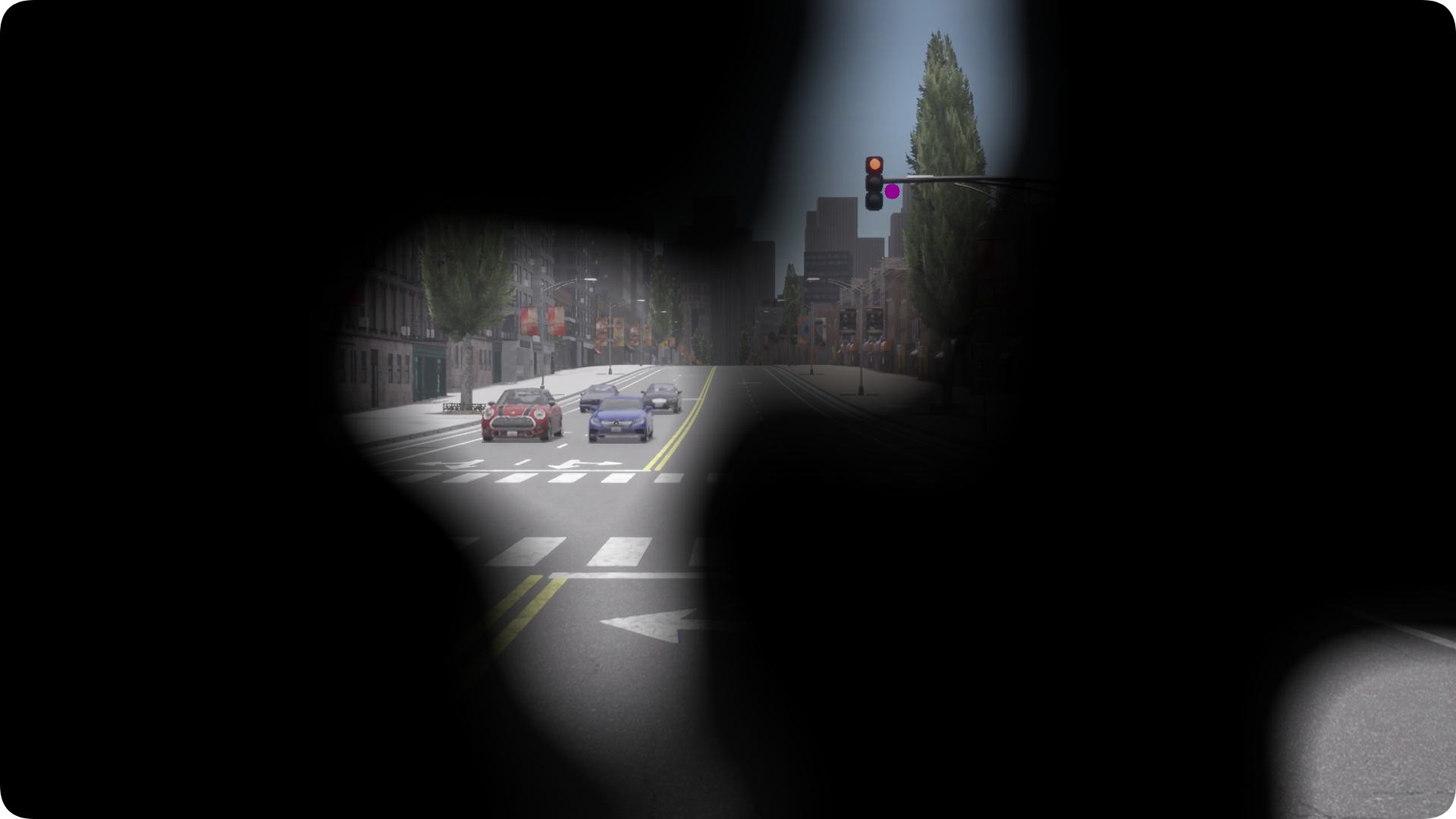}
            }     
            \subfloat{%
                \includegraphics[width=0.24\textwidth]{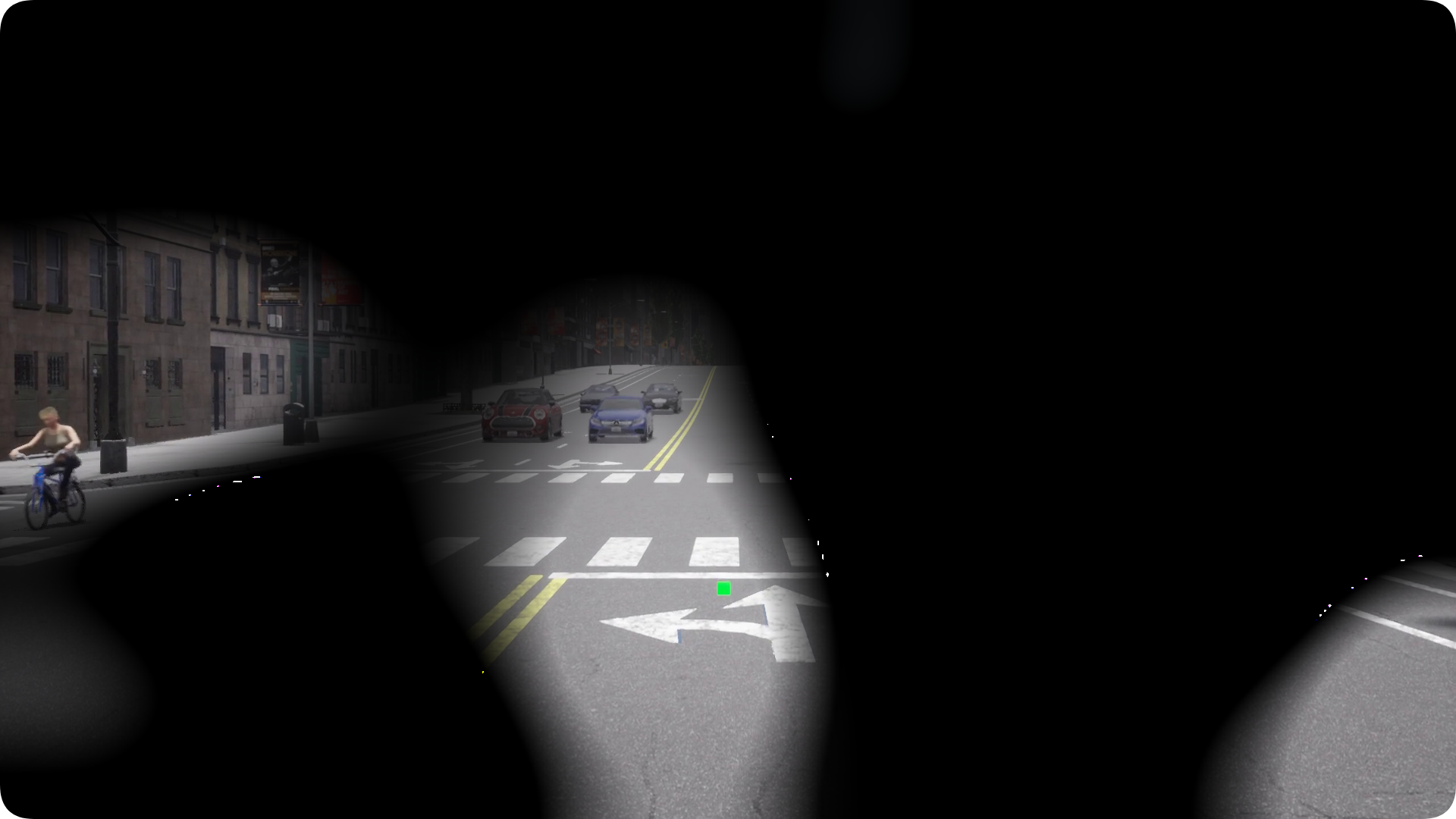}
            }   

            \setcounter{subfigure}{0}
            \subfloat[Image observation]{%
                \includegraphics[width=0.24\textwidth]{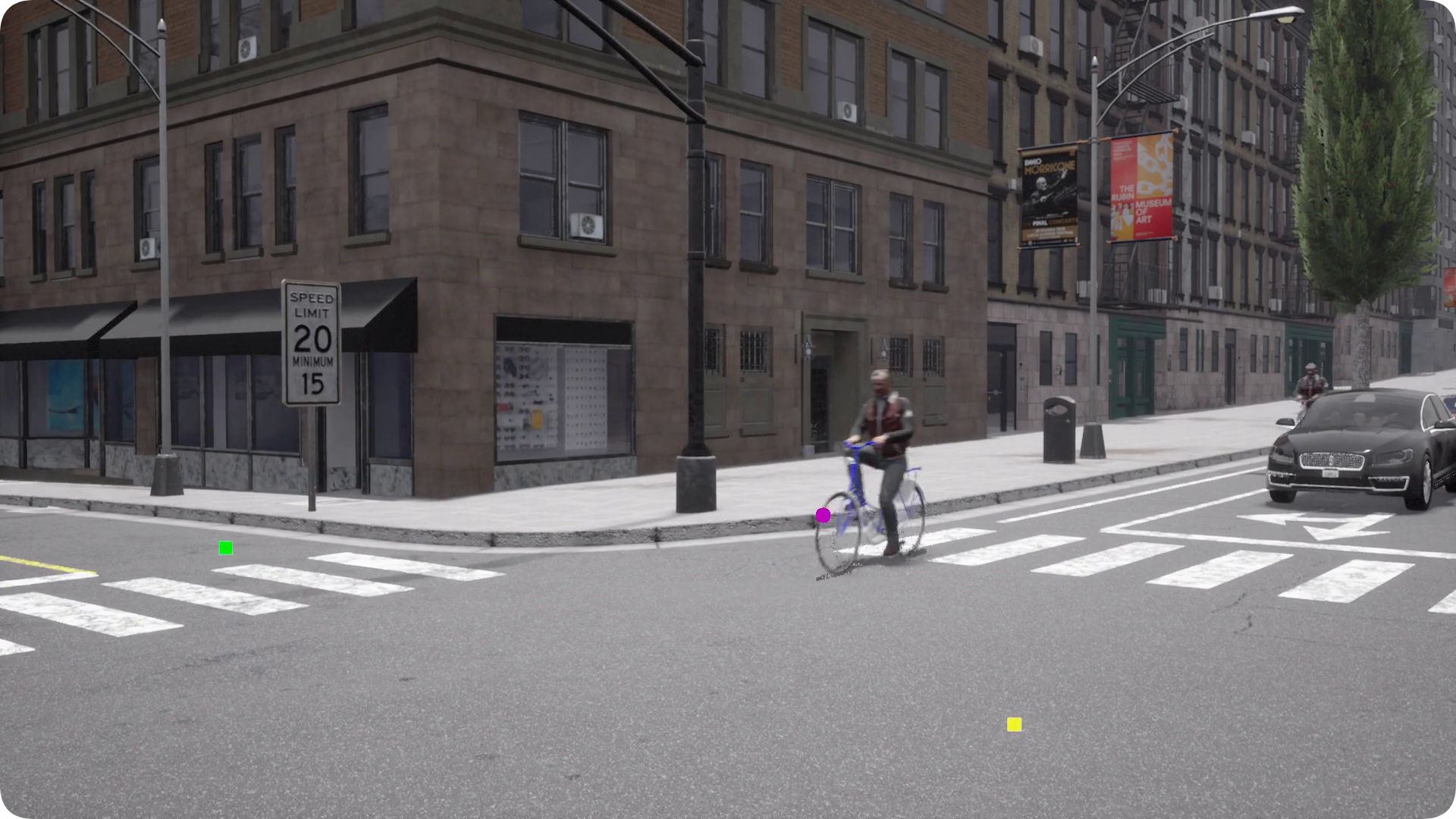}
            }
            \subfloat[Human gaze mask]{%
                \includegraphics[width=0.24\textwidth]{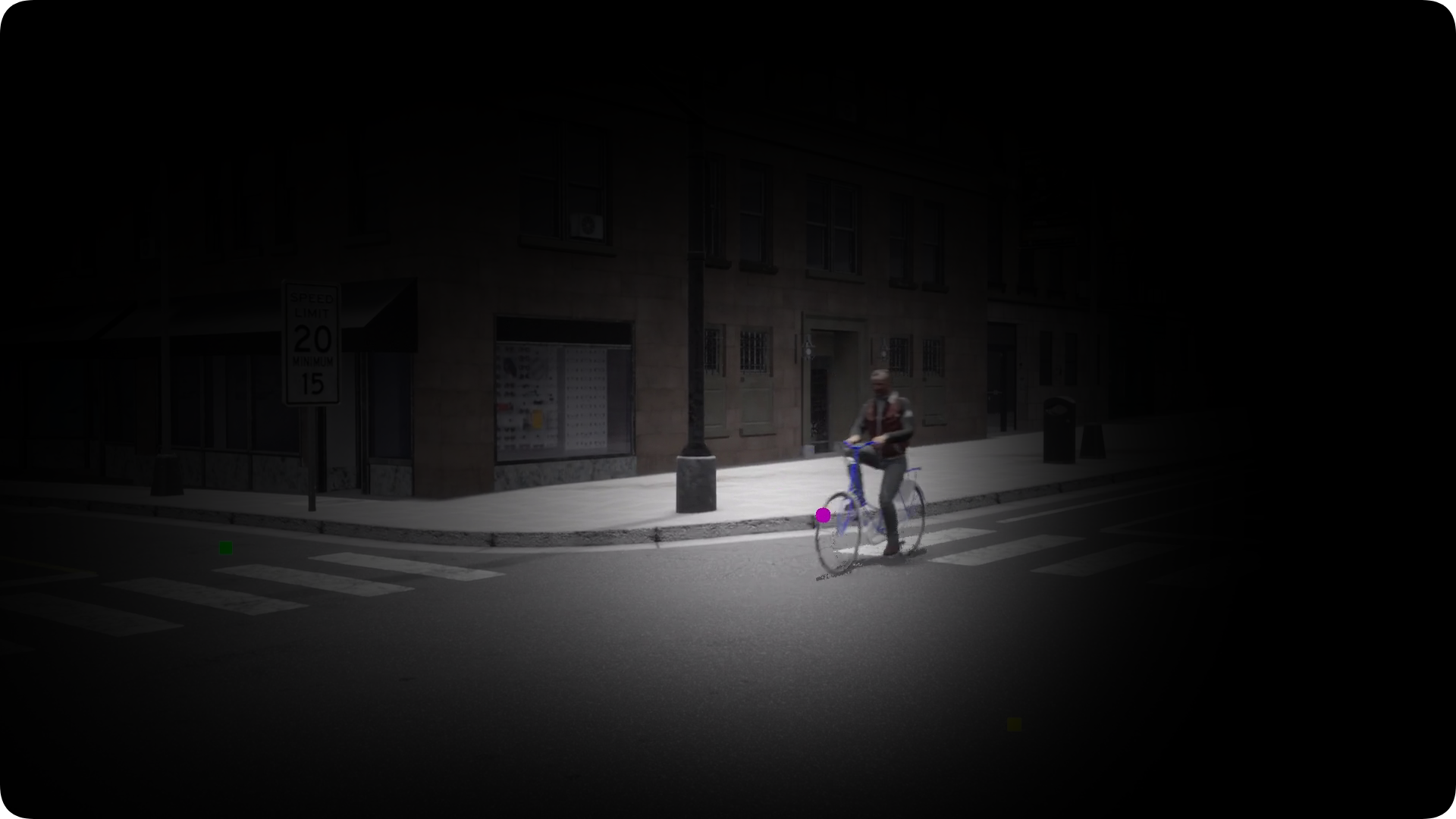}
            }
            \subfloat[GABRIL attention map]{%
                \includegraphics[width=0.24\textwidth]{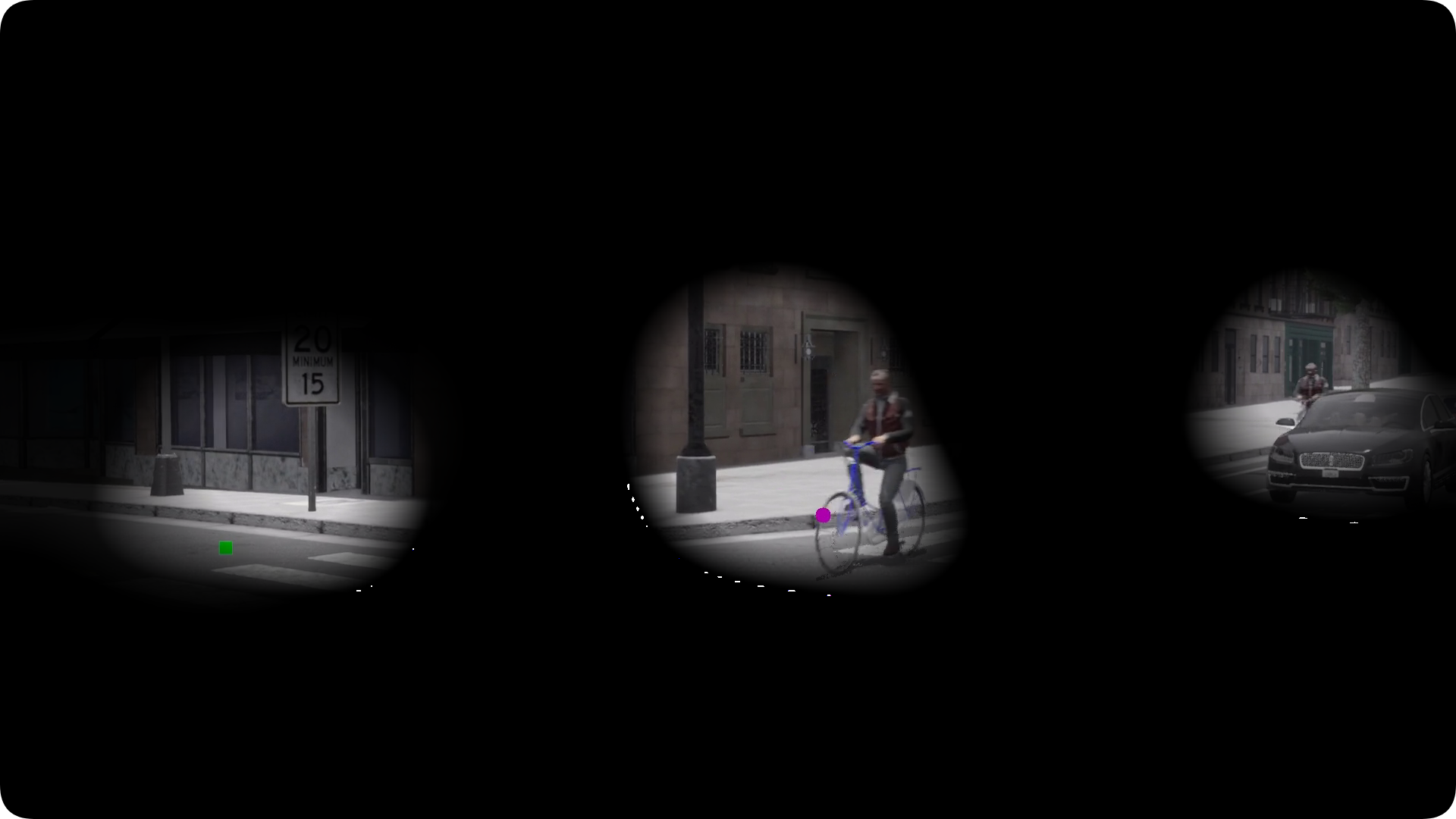}
            }     
            \subfloat[BC attention map]{%
                \includegraphics[width=0.24\textwidth]{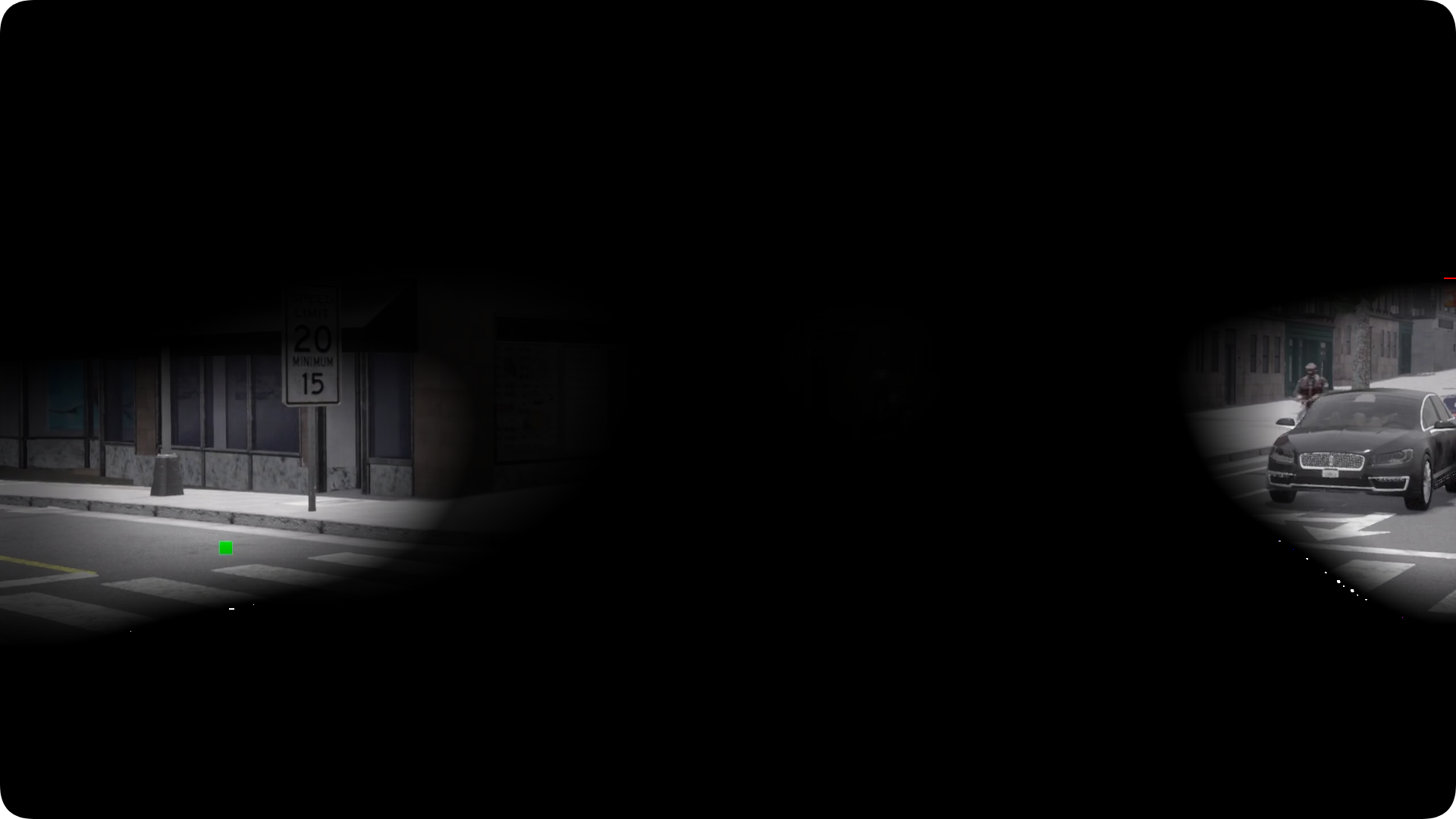}
            }          
    \caption{Comparing the interpretability of GaBRIL against a standard BC agent. The top row shows a situation in which the agent is waiting at the stoplight to turn left at an intersection. The bottom row shows a few moments later when the agent must yield to an incoming bicycle before completing the turn. The attention maps are constructed by rescaling the activation map of the last CNN layer and applying a Gaussian smoothing filter. As shown here, GABRIL actively directs its attention to causally relevant factors in the environment, whereas the regular BC agent is not as interpretable as GABRIL.}
    \label{fig_explainability}
    \vspace{-5mm}
\end{figure*}

\noindent\textbf{CARLA Results.}
Fig.~\ref{fig:carla} shows the results in the CARLA environment. We train a single model for each baseline using all 200 demonstrations. We exclude OREO due to its low training speed. For evaluations, we randomly sample 40 routes among the routes seen in the dataset with a different seed, as well as unseen routes with different combinations of weather and lighting conditions. The results indicate that GABRIL improves BC in the seen routes by $8.2\%$ and by $8.5\%$ in the unseen routes, which is the best among all other baselines. We also note that GABRIL trains much faster than the next best method, Contrastive, as it needs only a single pass through the encoder.

\noindent\textbf{Explainability.}
Since our model is equipped with an innate gaze-predictor, one can use the resulting activation map in real-time to visualize what the agent is attending to when making predictions \cite{oreo}. Fig.~\ref{fig_explainability} presents two instances of using this capability. For example, the agent trained by our method clearly considers the bike, the cars from the opposite lane, and the destination road when turning left at an intersection. However, in regular BC the activations are not highly interpretable or might miss essential elements in the environment, such as traffic lights or pedestrians. This capability is significantly crucial for real-world applications.

\begin{table*}[th!]
\vspace{+4mm}
\caption{
Comparing the performance of different methods in the \textit{normal} Atari environment.
}
\vspace{-1mm}

\label{tbl:main_normal}
\centering

\setlength{\tabcolsep}{4.5pt}
{\renewcommand{\arraystretch}{1.0}
\begin{tabular}{lccccccccccc}

\toprule

 & \multicolumn{1}{l}{} & \multicolumn{2}{c}{Architecture-based} & \multicolumn{4}{c}{Regularization-based} & \multicolumn{4}{c}{Dropout-based} \\
 
\cmidrule(lr){3-4}\cmidrule(lr){5-8} \cmidrule(lr){9-12}
 
Environment & BC & ViSaRL & AGIL & Contrastive & CGL & GRIL & GABRIL & GMD & \begin{tabular}[c]{@{}c@{}}GMD\\ + GABRIL\end{tabular} & OREO & \begin{tabular}[c]{@{}c@{}}OREO\\ + GABRIL\end{tabular} \\

\midrule

Alien & 1395.1 & 1484.8 & 1518.1 & 1459.5 & 1485.9 & 1609.7 & 1576.2 & 1999.9 & \textbf{2055.0} & 1614.1 & 1765.5 \\ 

Assault & 985.0 & 940.3 & 997.8 & 999.0 & 985.8 & 1002.6 & 951.0 & 937.5 & 1008.0 & 1020.4 & \textbf{1087.8} \\ 

Asterix & 645.4 & 482.5 & 685.4 & 622.8 & 601.7 & 649.4 & 701.9 & 717.8 & \textbf{782.4} & 709.2 & 708.9 \\ 

Breakout & 6.8 & 1.6 & 6.2 & 5.9 & 6.7 & 9.8 & 8.2 & 8.6 & \textbf{10.6} & 6.2 & 7.9 \\ 

ChopperCommand & 3046.0 & 2226.8 & 2470.0 & 3057.5 & 3106.8 & 2883.2 & 3313.2 & 2839.6 & \textbf{3690.8} & 3242.7 & 3297.0 \\ 

DemonAttack & 733.9 & 553.8 & 830.0 & 786.4 & 733.3 & 777.3 & 1122.7 & 491.4 & 1166.1 & 1080.3 & \textbf{1562.9} \\ 

Enduro & 318.7 & 299.3 & 323.2 & 319.7 & 319.1 & 313.7 & 315.3 & 330.1 & 321.3 & \textbf{337.1} & 305.5 \\ 

Freeway & 22.43 & 23.16 & \textbf{24.08} & 22.41 & 22.46 & 22.60 & 22.59 & 22.88 & 22.85 & 22.48 & 22.94 \\ 

Frostbite & 1652.0 & 1761.6 & 1901.8 & 1569.0 & 1575.9 & 1668.8 & 1887.7 & \textbf{2416.3} & 2008.8 & 1970.9 & 2013.4 \\ 

MsPacman & 1958.9 & 2419.1 & 2275.1 & 1813.1 & 2049.6 & 2030.9 & 1995.4 & \textbf{2505.3} & 2298.8 & 2317.4 & 2275.2 \\ 

Phoenix & 2897.6 & 1838.2 & 2721.3 & 2931.4 & 2854.2 & 2510.5 & 3781.4 & 2616.3 & 4464.8 & 4404.0 & \textbf{4492.5} \\ 

Qbert & 8645.8 & 7929.6 & 8393.2 & 9145.9 & 8515.0 & 8901.2 & 9305.9 & 9258.4 & 7238.8 & 10084.6 & \textbf{11022.7} \\ 

RoadRunner & 13118.4 & 12579.8 & 14129.8 & 15040.4 & 15590.8 & 13865.2 & 12759.2 & 13597.5 & 13222.4 & \textbf{15621.9} & 13128.0 \\ 

Seaquest & 506.3 & 601.0 & 582.0 & 555.4 & 526.1 & 591.8 & 674.9 & 560.4 & \textbf{690.7} & 463.9 & 599.7 \\ 

UpNDown & 5661.5 & 4407.1 & 5461.4 & 5853.0 & 5714.9 & 5610.2 & 6258.1 & 6527.8 & \textbf{6543.7} & 5531.2 & 6029.7 \\

\midrule

Mean ABC & 0.0\%  & -11.8\%  & 3.5\%  & 1.3\%  & 1.4\%  & 5.2\%  & 13.0\%  & 9.6\%  & \textbf{22.7\%}  & 13.0\%  & 21.9\%  \\

Median ABC & 0.0\%  & -6.1\%  & 6.2\%  & 1.2\%  & 0.1\%  & 1.8\%  & 8.8\%  & 7.1\%  & \textbf{21.2\%}  & 9.9\%  & 16.1\%  \\

\bottomrule

\end{tabular}
}
\end{table*}
\begin{table*}[th!]
\caption{
Comparing the performance of different methods in the \textit{confounded} Atari environment.
}
\vspace{-1mm}

\label{tbl:main_confounded}
\centering

\setlength{\tabcolsep}{5pt}
{\renewcommand{\arraystretch}{1.01}
\begin{tabular}{lccccccccccc}

\toprule

 & \multicolumn{1}{l}{} & \multicolumn{2}{c}{Architecture-based} & \multicolumn{4}{c}{Regularization-based} & \multicolumn{4}{c}{Dropout-based} \\
 
\cmidrule(lr){3-4}\cmidrule(lr){5-8} \cmidrule(lr){9-12}
 
Environment & BC & ViSaRL & AGIL & Contrastive & CGL & GRIL & GABRIL & GMD & \begin{tabular}[c]{@{}c@{}}GMD\\ + GABRIL\end{tabular} & OREO & \begin{tabular}[c]{@{}c@{}}OREO\\ + GABRIL\end{tabular} \\

\midrule

Alien & 1002.2 & 1269.2 & 1408.7 & 905.7 & 916.8 & 1088.9 & 1148.8 & \textbf{1577.5} & 1360.8 & 830.8 & 1367.8 \\ 

Assault & 730.4 & 721.3 & 723.7 & 705.5 & 720.6 & 728.3 & 773.0 & 681.7 & 722.8 & 707.6 & \textbf{1051.9} \\ 

Asterix & 514.1 & 503.3 & 656.7 & 493.7 & 486.8 & 506.0 & \textbf{689.9} & 595.1 & 598.1 & 609.7 & 616.4 \\ 

Breakout & 5.0 & 1.3 & 3.3 & 5.0 & 4.8 & 8.6 & 7.9 & 5.1 & \textbf{10.0} & 3.5 & 7.1 \\ 

ChopperCommand & 2270.0 & 2242.5 & 2547.0 & 2511.0 & 2337.5 & 2489.0 & 2759.0 & 2768.3 & \textbf{2933.0} & 2577.0 & 2818.0 \\ 

DemonAttack & 263.2 & 301.5 & 318.3 & 309.0 & 251.7 & 270.2 & 436.7 & 182.4 & 313.4 & 492.1 & \textbf{578.0} \\ 

Enduro & 291.0 & 262.3 & 294.1 & 289.2 & 282.6 & 252.8 & 306.8 & 295.0 & 303.4 & 308.5 & \textbf{311.9} \\ 

Freeway & 22.14 & 22.65 & 22.98 & 22.26 & 22.66 & 22.67 & 22.77 & 22.71 & 22.96 & 22.82 & \textbf{23.20} \\ 

Frostbite & 990.7 & 1383.5 & 1266.9 & 939.1 & 1007.8 & 1311.3 & 1601.9 & \textbf{1675.5} & 1486.8 & 1116.4 & 1295.1 \\ 

MsPacman & 1435.1 & 1587.1 & 1443.8 & 1320.5 & 1360.6 & 1476.8 & 1637.3 & \textbf{1905.5} & 1811.1 & 1520.3 & 1621.1 \\ 

Phoenix & 1638.4 & 1061.2 & 1401.4 & 1539.0 & 1662.1 & 1396.1 & 2006.9 & 992.4 & 2080.3 & 2022.0 & \textbf{2191.7} \\ 

Qbert & 3597.1 & 3979.2 & 3434.7 & 3709.6 & 4961.0 & 4536.9 & \textbf{5839.8} & 4970.5 & 5679.6 & 4507.9 & 4509.8 \\ 

RoadRunner & 9177.3 & 7452.8 & 10843.6 & 10133.7 & 8630.4 & 11340.8 & 9165.7 & \textbf{15600.0} & 14909.0 & 9818.8 & 8160.0 \\ 

Seaquest & 405.8 & 472.4 & 464.0 & 415.8 & 406.9 & 520.3 & \textbf{567.7} & 468.3 & 557.8 & 388.9 & 496.7 \\ 

UpNDown & 4270.2 & 4255.6 & 4578.6 & 4312.4 & 4438.1 & 4897.6 & 4273.8 & 5089.4 & 5300.0 & 5246.0 & \textbf{5430.5} \\ 

\midrule

Mean ABC & 0.0\%  & -1.4\%  & 8.0\%  & 0.6\%  & 0.8\%  & 12.8\%  & 27.1\%  & 17.9\%  & \textbf{32.7\%}  & 11.3\%  & 29.3\%  \\
Median ABC & 0.0\%  & -0.3\%  & 7.2\%  & -0.3\%  & -1.3\%  & 8.6\%  & 21.5\%  & 15.8\%  & \textbf{27.0\%}  & 7.0\%  & 25.4\%  \\

\bottomrule

\end{tabular}
}
\vspace{-5mm}
\end{table*}

\section{Limitations and Future Work} 

In this paper, we introduced GABRIL, a regularization method for mitigating causal confusion in imitation learning. However, our approach has several limitations. First, causal confusion arises not only across the spatial dimension but also over time, a phenomenon known as the copycat problem \cite{resolve_copycat, fighting_copycat}. This issue is particularly pronounced in environments where expert actions exhibit high temporal correlation. Notably, our method is not designed to address this type of causal confusion. Future research should explore the use of gaze to mitigate these temporal dependencies.

Additionally, collecting gaze data can be challenging or noisy in real-world robotic settings where the expert operates in an open environment. Future work should focus on improving data efficiency to better integrate gaze information into imitation learning.

\section*{Acknowledgments}
This work was sponsored in part by DEVCOM Army Research Laboratory under Cooperative Agreement Number W911NF-19-S-0001. Yutai Zhou was partially supported by a fellowship from USC - Capital One
Center for Responsible AI and Decision Making in Finance (CREDIF).

\balance
\printbibliography

@inproceedings{biswas2024,
author = {Biswas, Abhijat and Pardhi, Badal Arun and Chuck, Caleb and Holtz, Jarrett and Niekum, Scott and Admoni, Henny and Allievi, Alessandro},
title = {Gaze Supervision for Mitigating Causal Confusion in Driving Agents},
year = {2024},
publisher = {International Foundation for Autonomous Agents and Multiagent Systems},
address = {Richland, SC},
booktitle = {Proceedings of the 23rd International Conference on Autonomous Agents and Multiagent Systems},
pages = {2159–2161},
series = {AAMAS '24}
}

@inproceedings{visarl,
 author = {Liang, Anthony and Thomason, Jesse and Bıyık, Erdem},
 title = {ViSaRL: Visual Reinforcement Learning Guided by Human Saliency},
 booktitle = {International Conference on Intelligent Robots and Systems (IROS)},
 year = {2024}
}

@inproceedings{oreo,
 author = {Park, Jongjin and Seo, Younggyo and Liu, Chang and Zhao, Li and Qin, Tao and Shin, Jinwoo and Liu, Tie-Yan},
 booktitle = {Advances in Neural Information Processing Systems},
 pages = {3029--3042},
 publisher = {Curran Associates, Inc.},
 title = {Object-Aware Regularization for Addressing Causal Confusion in Imitation Learning},
 volume = {34},
 year = {2021}
}

@inproceedings{AGIL,
author = {Zhang, Ruohan and Liu, Zhuode and Zhang, Luxin and Whritner, Jake A. and Muller, Karl S. and Hayhoe, Mary M. and Ballard, Dana H.},
title = {AGIL: Learning Attention from Human for Visuomotor Tasks},
year = {2018},
publisher = {Springer-Verlag},
booktitle = {Computer Vision – ECCV 2018: 15th European Conference, Munich, Germany, September 8-14, 2018, Proceedings, Part XI},
pages = {692–707},
}

@inproceedings{CGL,
author = {Saran, Akanksha and Zhang, Ruohan and Short, Elaine S. and Niekum, Scott},
title = {Efficiently Guiding Imitation Learning Agents with Human Gaze},
year = {2021},
publisher = {International Foundation for Autonomous Agents and Multiagent Systems},
address = {Richland, SC},
booktitle = {Proceedings of the 20th International Conference on Autonomous Agents and MultiAgent Systems},
pages = {1109–1117},
numpages = {9},
}

@article{GRIL,
  title={Gaze-Informed Multi-Objective Imitation Learning from Human Demonstrations},
  author={Bera, Ritwik and Goecks, Vinicius G and Gremillion, Gregory M and Lawhern, Vernon J and Valasek, John and Waytowich, Nicholas R},
  journal={arXiv preprint arXiv:2102.13008},
  year={2021}
}

@inproceedings{zagoruyko2017paying,
title={Paying More Attention to Attention: Improving the Performance of Convolutional Neural Networks via Attention Transfer},
author={Sergey Zagoruyko and Nikos Komodakis},
booktitle={International Conference on Learning Representations},
year={2017},
}

@inproceedings{Laskin2020Reinforcement,
 author = {Laskin, Misha and Lee, Kimin and Stooke, Adam and Pinto, Lerrel and Abbeel, Pieter and Srinivas, Aravind},
 booktitle = {Advances in Neural Information Processing Systems},
 pages = {19884--19895},
 publisher = {Curran Associates, Inc.},
 title = {Reinforcement Learning with Augmented Data},
 volume = {33},
 year = {2020}
}

@ARTICLE{gaze_robot_manipulation_1,
  author={Kim, Heecheol and Ohmura, Yoshiyuki and Kuniyoshi, Yasuo},
  journal={IEEE Robotics and Automation Letters}, 
  title={Using Human Gaze to Improve Robustness Against Irrelevant Objects in Robot Manipulation Tasks}, 
  year={2020},
  volume={5},
  number={3},
  pages={4415--4422},
}

@INPROCEEDINGS{gaze_robot_manipulation_2,
  author={Kim, Heecheol and Ohmura, Yoshiyuki and Kuniyoshi, Yasuo},
  booktitle={2022 International Conference on Robotics and Automation (ICRA)}, 
  title={Memory-based gaze prediction in deep imitation learning for robot manipulation}, 
  year={2022},
  volume={},
  number={},
  pages={2427--2433},
}

@inproceedings{Causal_Confusion_in_IL,
 author = {de Haan, Pim and Jayaraman, Dinesh and Levine, Sergey},
 booktitle = {Advances in Neural Information Processing Systems},
 pages = {},
 publisher = {Curran Associates, Inc.},
 title = {Causal Confusion in Imitation Learning},
 volume = {32},
 year = {2019}
}

@article{selective_gaze,
  title={Selective eye-gaze augmentation to enhance imitation learning in Atari games},
  author={Thammineni, Chaitanya and Manjunatha, Hemanth and Esfahani, Ehsan T},
  journal={Neural Computing and Applications},
  volume={35},
  number={32},
  pages={23401--23410},
  year={2023},
  publisher={Springer}
}

@INPROCEEDINGS{GMD,
  author={Chen, Yuying and Liu, Congcong and Tai, Lei and Liu, Ming and Shi, Bertram E.},
  booktitle={2019 IEEE/RSJ International Conference on Intelligent Robots and Systems (IROS)}, 
  title={Gaze Training by Modulated Dropout Improves Imitation Learning}, 
  year={2019},
  volume={},
  number={},
  pages={7756--7761},
}

@InProceedings{SEMI_cc,
author="Zhang, Huanghui
and Zheng, Zhi",
title="Sequential Masking Imitation Learning for Handling Causal Confusion in Autonomous Driving",
booktitle="Advanced Computational Intelligence and Intelligent Informatics",
year="2024",
publisher="Springer Nature Singapore",
pages="200--214",
}

@article{causal_deconfounding_drl,
title = {Causal deconfounding deep reinforcement learning for mobile robot motion planning},
journal = {Knowledge-Based Systems},
volume = {303},
pages = {112406},
year = {2024},
author = {Wenbing Tang and Fenghua Wu and Shang-wei Lin and Zuohua Ding and Jing Liu and Yang Liu and Jifeng He},
}

@inproceedings{bench2drive,
  title={Bench2Drive: Towards Multi-Ability Benchmarking of Closed-Loop End-To-End Autonomous Driving},
  author={Xiaosong Jia and Zhenjie Yang and Qifeng Li and Zhiyuan Zhang and Junchi Yan},
  booktitle={NeurIPS 2024 Datasets and Benchmarks Track},
  year={2024}
}

@inproceedings{seeing_is_not_believing,
 author = {Ding, Wenhao and Shi, Laixi and Chi, Yuejie and ZHAO, DING},
 booktitle = {Advances in Neural Information Processing Systems},
 pages = {66328--66363},
 publisher = {Curran Associates, Inc.},
 title = {Seeing is not Believing: Robust Reinforcement Learning against Spurious Correlation},
 volume = {36},
 year = {2023}
}

@inproceedings{CARLA,
  title = { {CARLA}: {An} Open Urban Driving Simulator},
  author = {Alexey Dosovitskiy and German Ros and Felipe Codevilla and Antonio Lopez and Vladlen Koltun},
  booktitle = {Proceedings of the 1st Annual Conference on Robot Learning},
  pages = {1--16},
  year = {2017}
}

@article{fixation_saccade,
title = {Eye tracking algorithms, techniques, tools, and applications with an emphasis on machine learning and Internet of Things technologies},
journal = {Expert Systems with Applications},
volume = {166},
pages = {114037},
year = {2021},
author = {Ahmad F. Klaib and Nawaf O. Alsrehin and Wasen Y. Melhem and Haneen O. Bashtawi and Aws A. Magableh},
}

@inproceedings{resolve_copycat,
  title={Resolving copycat problems in visual imitation learning via residual action prediction},
  author={Chuang, Chia-Chi and Yang, Donglin and Wen, Chuan and Gao, Yang},
  booktitle={European Conference on Computer Vision},
  pages={392--409},
  year={2022},
  organization={Springer}
}

@article{gymnasium,
  title={Gymnasium: A Standard Interface for Reinforcement Learning Environments},
  author={Towers, Mark and Kwiatkowski, Ariel and Terry, Jordan and Balis, John U and De Cola, Gianluca and Deleu, Tristan and Goul{\~a}o, Manuel and Kallinteris, Andreas and Krimmel, Markus and KG, Arjun and others},
  journal={arXiv preprint arXiv:2407.17032},
  year={2024}
}

@article{fighting_copycat,
  title={Fighting copycat agents in behavioral cloning from observation histories},
  author={Wen, Chuan and Lin, Jierui and Darrell, Trevor and Jayaraman, Dinesh and Gao, Yang},
  journal={Advances in Neural Information Processing Systems},
  volume={33},
  pages={2564--2575},
  year={2020}
}

@article{survey_end_to_end_autonomous,
  title={End-to-end Autonomous Driving: Challenges and Frontiers},
  author={Chen, Li and Wu, Penghao and Chitta, Kashyap and Jaeger, Bernhard and Geiger, Andreas and Li, Hongyang},
  journal={IEEE Transactions on Pattern Analysis and Machine Intelligence},
  year={2024}
}

@ARTICLE{Bylinskii2019WhatDo,
  author={Bylinskii, Zoya and Judd, Tilke and Oliva, Aude and Torralba, Antonio and Durand, Frédo},
  journal={IEEE Transactions on Pattern Analysis and Machine Intelligence}, 
  title={What Do Different Evaluation Metrics Tell Us About Saliency Models?}, 
  year={2019},
  volume={41},
  number={3},
  pages={740-757},
}

@article{hollingworth2008understanding,
  title={Understanding the function of visual short-term memory: transsaccadic memory, object correspondence, and gaze correction.},
  author={Hollingworth, Andrew and Richard, Ashleigh M and Luck, Steven J},
  journal={Journal of Experimental Psychology: General},
  volume={137},
  number={1},
  pages={163},
  year={2008},
  publisher={American Psychological Association}
}

@article{cowan2001magical,
  title={The magical number 4 in short-term memory: A reconsideration of mental storage capacity},
  author={Cowan, Nelson},
  journal={Behavioral and brain sciences},
  volume={24},
  number={1},
  pages={87--114},
  year={2001},
  publisher={Cambridge University Press}
}

@inproceedings{ALVINN,
 author = {Pomerleau, Dean A.},
 booktitle = {Advances in Neural Information Processing Systems},
 publisher = {Morgan--Kaufmann},
 title = {ALVINN: An Autonomous Land Vehicle in a Neural Network},
 volume = {1},
 year = {1988}
}

@misc{leaderboard_v2,
    title  = {Leaderboard 2.0},
    author = {},
    url    = {https://leaderboard.carla.org/},
    urldate   = {2025-25-02},
    note = {Accessed: 2025-25-02}
}

@inproceedings{vqvae,
 author = {van den Oord, Aaron and Vinyals, Oriol and kavukcuoglu, koray},
 booktitle = {Advances in Neural Information Processing Systems},
 pages = {},
 publisher = {Curran Associates, Inc.},
 title = {Neural Discrete Representation Learning},
 volume = {30},
 year = {2017}
}

@inproceedings{clrl,
  title={Causally regularized learning with agnostic data selection bias},
  author={Shen, Zheyan and Cui, Peng and Kuang, Kun and Li, Bo and Chen, Peixuan},
  booktitle={Proceedings of the 26th ACM international conference on Multimedia},
  pages={411--419},
  year={2018}
}

@article{Darby2021Attention,
author = {Darby, Kevin P. and Deng, Sophia W. and Walther, Dirk B. and Sloutsky, Vladimir M.},
title = {The Development of Attention to Objects and Scenes: From Object-Biased to Unbiased},
journal = {Child Development},
volume = {92},
number = {3},
pages = {1173-1186},
year = {2021}
}

@ARTICLE{Itti1998Saliency,
  author={Itti, L. and Koch, C. and Niebur, E.},
  journal={IEEE Transactions on Pattern Analysis and Machine Intelligence}, 
  title={A model of saliency-based visual attention for rapid scene analysis}, 
  year={1998},
  volume={20},
  number={11},
  pages={1254-1259},
}

@book{peters2017elements,
  title={Elements of causal inference: foundations and learning algorithms},
  author={Peters, Jonas and Janzing, Dominik and Sch{\"o}lkopf, Bernhard},
  year={2017},
  publisher={The MIT Press}
}

@article{geirhos2020shortcut,
  title={Shortcut learning in deep neural networks},
  author={Geirhos, Robert and Jacobsen, J{\"o}rn-Henrik and Michaelis, Claudio and Zemel, Richard and Brendel, Wieland and Bethge, Matthias and Wichmann, Felix A},
  journal={Nature Machine Intelligence},
  volume={2},
  number={11},
  pages={665--673},
  year={2020},
  publisher={Nature Publishing Group UK London}
}

@article{hinton2015distilling,
  title={Distilling the knowledge in a neural network},
  author={Hinton, Geoffrey and Vinyals, Oriol and Dean, Jeff},
  journal={arXiv preprint arXiv:1503.02531},
  year={2015}
}

@inproceedings{zhang2020atari,
  title={Atari-head: Atari human eye-tracking and demonstration dataset},
  author={Zhang, Ruohan and Walshe, Calen and Liu, Zhuode and Guan, Lin and Muller, Karl and Whritner, Jake and Zhang, Luxin and Hayhoe, Mary and Ballard, Dana},
  booktitle={Proceedings of the AAAI conference on artificial intelligence},
  volume={34},
  number={04},
  pages={6811--6820},
  year={2020}
}

\end{document}